\documentclass[smallcondensed,natbib]{svjour3}
\pdfoutput=1

\usepackage[latin1]{inputenc}
\usepackage{xspace}
\usepackage{multirow}
\usepackage{url}
\usepackage{xcolor}
\usepackage{ctable}
\usepackage{colortbl}
\usepackage{graphicx}
\usepackage{subfigure}

\newcommand{\ie}{\emph{i.e.}\/\xspace}
\newcommand{\mx}{\hbox{\textbf{x}}\xspace}

\newcommand{\lp}{{\emph {LP}}\/\xspace}

\newcommand{\lbg}{\hbox{{\textit{General}}$_B$}\xspace}

\newcommand{\bibtex}{\textit{Bibtex}\xspace}
\newcommand{\corel}{\textit{Corel5k}\xspace}
\newcommand{\emotions}{\textit{Emotions}\xspace}
\newcommand{\enron}{\textit{Enron}\xspace}
\newcommand{\genbase}{\textit{Genbase}\xspace}
\newcommand{\mediamill}{\textit{Mediamill}\xspace}
\newcommand{\medical}{\textit{Medical}\xspace}
\newcommand{\scene}{\textit{Scene}\xspace}
\newcommand{\slashdot}{\textit{Slashdot}\xspace}
\newcommand{\yeast}{\textit{Yeast}\xspace}

\newcommand{\subacc}{\hbox{\textit{Subset-Accuracy}}\xspace}
\newcommand{\hloss}{\hbox{\textit{Hamming-Loss}}\xspace}
\newcommand{\accuracy}{\hbox{\textit{Accuracy}}\xspace}
\newcommand{\fmeasure}{\hbox{\textit{F-Measure}}\xspace}

\newcommand{\acc}{\textit{Acc}\xspace}
\newcommand{\f}{\textit{F1}\xspace}
\newcommand{\fM}{\textit{F1$^M$}\xspace}
\newcommand{\fm}{\textit{F1$^\mu$}\xspace}
\newcommand{\hl}{\textit{HL}\xspace}
\newcommand{\pr}{\textit{Pr}\xspace}
\newcommand{\re}{\textit{Re}\xspace}
\newcommand{\sacc}{\textit{SAcc}\xspace}

\definecolor{lightgray}{gray}{0.85}

\journalname{}

\begin{document}

\title{Comparing published multi-label classifier performance measures  to the ones obtained by a simple multi-label baseline classifier\footnote{Acknowledgements: This research was supported by the S\~ao Paulo Research Foundation (FAPESP), grants  2008/06739-0, 2010/15992-0 and 2011/02393-4.}}


\author{Jean Metz \and
Newton Spola\^or   \and
Everton A. Cherman \and
Maria C. Monard
}

\authorrunning{Jean Metz et al.} 

\institute{Jean Metz \at
                Federal Technological University of Paran\'a  (UTFPR)\\
                P.O. Box 271, Zip Code 85884-000, Medianeira, PR, Brazil \\
              Tel.: +55-45-32408119\\
              Fax: +55-45-32408101\\
              \email{jeanmetz@utfpr.edu.br}           
           \and
            Newton Spola\^or \and Everton A. Cherman \and 
Maria C. Monard \at
              Institute of Mathematics and Computer Science (ICMC)\\
                University of S\~ao Paulo (USP) \\
                P.O. Box 668, Zip Code 13561-970, S\~ao Carlos, SP, Brazil \\
              Tel.: +55-16-33739700 \\
              Fax: +55-16-33712238 \\
              \email{ \{newtonsp,echerman,mcmonard\}@icmc.usp.br}
}

\date{Received: date / Accepted: date}

\maketitle

\begin{abstract}
In supervised learning, simple baseline classifiers can be constructed by only looking at the class, \ie, ignoring any other information from the dataset.  The single-label learning community frequently uses as a reference the one which always predicts the majority class. Although a classifier might perform worse than this simple baseline classifier, this behaviour requires a special explanation. Aiming to motivate the community to compare experimental results with the ones provided by a multi-label baseline classifier, calling the attention about the need of special explanations related to classifiers which perform worse than the baseline, in this work we propose the use of \lbg, a multi-label baseline classifier. \lbg was evaluated in contrast to results published in the literature which were carefully selected using a systematic review process. It was found that a considerable number of published results on 10 frequently used datasets are worse than or equal to the ones obtained by \lbg, and for one dataset it reaches up to 43\% of the dataset published results. Moreover, although a simple baseline classifier was not considered in these publications, it was observed that even for very poor results no special explanations were provided in most of them. We hope that the findings of this work would encourage the multi-label community to consider the idea of using a simple baseline classifier, such that further explanations are provided when a classifiers performs worse than a baseline.

\keywords{machine learning \and 
multi-label classification \and 
multi-label baseline classifier \and
systematic review}

\end{abstract}

\section{Introduction}
Given a set of examples (instances)
characterized by the value of attributes and the class of the example, the aim of
supervised learning algorithms is to construct a classifier which is
able to assign new examples to the  class they  belong to. To this
end, a great deal of learning algorithms have been proposed.  The
constructed classifiers are usually compared over a variety of datasets using
various evaluation measures proposed in the literature. Most results
are  averages over a number of runs, where each  run involves
splitting the dataset into disjoint training and test sets, and
the test  set is used to estimate several evaluation measures  of
the classifier generated using the corresponding training set.
Afterwards, it is important to statistically verify the hypothesis
of improved performance  (or not) of the learning algorithm~\citep{Demsar2006}.
However, we consider that the evaluation measures of the classifier
constructed by a learning algorithm should also be compared with the
ones obtained by a simple baseline classifier, as it is actually done by
most of the single-label learning community. This way, case any of these 
measures are worse, it would encourage the community to provide 
additional explanation of this fact.

In  \textit{single-label learning}, each example in the dataset
 is associated with only one class, which can assume several values.
The task is called \textit{binary classification} if there are only
two possible class values (Yes/No), and \textit{multi-class
classification} when the number of class values is greater than
two~\citep{Alpaydin04Introduction}.

For single-label learning, a \textit{simple baseline classifier} is the one
constructed by only taking into account the class values, \ie, it does not 
consider the attributes that describe the examples in the dataset.
Having only this information, and due to the fact that the classification of a new
instance has only two possible outcomes, correct or incorrect,  the
best it can do is to output a classifier that always predicts the
most frequently occurring class value  in the dataset. Before the
single-label learning community started to pay attention to this
very simple baseline classifier, many evaluation measures worse than or
equal to this baseline classifier had been published in the scientific
literature, without special explanations.

In~\citep{Holte93} an experimental comparison  involving  16 commonly used single-label datasets is carried out, where the error rate of the proposed algorithms are compared to the error rate of several learning systems reported in the literature. However, although the datasets used are not highly skewed, some of these reported  results fail to improve the error rate of the simple baseline classifier. For example, considering  two  of these datasets, Breast Cancer and Hepatitis, from the collection distributed by the University of California at Irvine~\citep{Bache2013}, 33 and 75 error rates are compiled respectively. For the dataset Hepatitis 8 out of 33 (more than 24\%) reported error rates are worse than or equal to the simple baseline classifier, while for the dataset Breast Cancer the same happens for 29 out of 75      (more than 38\%) reported error rates. As the simple single-label baseline classifier is constructed by only looking at the class values, any learning algorithm which learns from non-skewed domains,  and also takes into account the dataset attribute values should be able to construct a classifier with smaller error rates.


Different to single-label learning, in \textit{multi-label learning} 
an example can belong to several classes simultaneously. The
main difference between multi-label learning and single-label
learning is that classes in multi-label learning are often
correlated, and the class values in single-label learning are
mutually exclusive. Due to the increasing number of  applications
where examples are annotated with more than one class, multi-label
learning has received increasing attention from the machine learning
community~\citep{Tsoumakas2010DataMining}.

However, finding a simple multi-label baseline
classifier by only looking at the multi-labels is not as straightforward 
as in single-label,  where the classification of a new
instance has only two possible outcomes, correct or incorrect, and the error rate is
often considered an important single objective to be achieved. This is not
the case in multi-label, as  the evaluation measures of a
multi-label classifier should also take into account
\textit{partially correct} classifications. To this end, many
criteria are proposed to evaluate the classification performance
from different perspectives.
In~\citep{DembczynskiWCH12}, the connection among
these criteria are established, showing that some of these criteria
are uncorrelated or even negatively correlated. In other words, some
loss functions are essentially conflicting. Thus, several
 multi-label evaluation measures have been proposed, highlighting
 different aspects of this important characteristic of multi-label learning.

 Motivated by the lack of simple multi-label baseline classifiers,
 in~\citep{MetzACM12} we propose a simple way to
 construct multi-label baseline classifiers for specific  multi-label evaluation measures.
Nevertheless, as a
 multi-label classifier which focuses on mini\-mi\-zing/ma\-xi\-mi\-zing one of
 these measures does not necessarily mi\-ni\-mi\-ze/ma\-xi\-mi\-ze the others, 
  we also proposed a unique simple baseline
 classifier, called \lbg, which does not focus on any one of these specific
 measures and can be used to determine all the multi-label evaluation measure
  baseline values of a classifier.

Although we do not claim that the proposed \lbg multi-label baseline
classifier should be the one to be used by the community
whenever classifiers evaluation measures are published, as other
baseline classifiers could be proposed in the future, we believe
that it is time to start a discussion related to this subject.
Aiming to motivate the community, in this work we consider published
experimental results which show that, similar to the single-label
research primordium, some of the published results fail to improve
on the ones obtained by our simple multi-label baseline classifier.

However, unlike~\citep{Holte93}, in which   results
reported on a dataset could also refer to  the classifier generated
by a learning algorithm using a slightly different dataset due to
pre-processing, such as filter feature selection or other
transformation, in this work we only used the results published in
papers reporting experimental results of classifiers which have been
constructed using publicy available \textit{identical} datasets. 
 Unfortunately, this constraint leaves out a great deal
of papers, such as many related to text categorization, a
typical multi-label problem, as most of the publicly available text datasets
are modified by the authors in different ways to obtain the final
dataset from which the classifier is generated and, in most cases,
this final dataset is not publicly  available. On the other hand,
this constraint enables anyone to reproduce the experiments described in these papers.

As there is a lack of reviews focusing on pieces of work which
report experimental results for multi-label learning, and the
systematic review process can be useful to identify related
publications in a wide, rigorous and replicable way~\citep{KitchenhamPBBTNL10}, we used this
process to identify  publications which report experimental results
for multi-label learning. We have gathered the data used
in this work from the selected publications which
answer the systematic review research question and do not fulfill any
of the exclusion criteria.

More specifically, in this work we report on several statistics of
various evaluation measure values, which were published and obtained
using the 10 datasets most frequently used in the selected papers.
These statistics show that 12.8\% of these published results are
worse than or equal to the ones obtained by our simple multi-label
baseline classifier \lbg. Moreover, this percentage is unevenly
distributed among the datasets. In the ``worst'' dataset, 43.0\% of such
results were reported, and in the ``best'' one only 0.6\%.
However, although a simple baseline classifier was not considered in these publications, it was observed that even for very poor results no special explanations were provided in most of these publications.

The remainder of this paper is organized as follows:
Section~\ref{sec:background} briefly describes multi-label learning
and the evaluation measures used in this work. Section~\ref{sec:lgb}
explains the simple baseline classifier \lbg. The systematic review
carried out  to select the papers from which we have gathered the data used
in this work is described in Section~\ref{sec:systematicreview}, and
 statistics of these published evaluation measure values are reported in
Section~\ref{sec:expdesign}. Section~\ref{sec:conclusion} presents
the conclusions and future work.

\section{Multi-label Classification and Evaluation Measures}
\label{sec:background}

Let $D$ be a  training set composed of $N$ examples $E_i =
(\mx_i,Y_i)$, $i = 1..N$. Each example $E_i$ is associated with a
feature vector $\mx_i = (x_{i1}, x_{i2}, \ldots, x_{iM})$ described
by $M$ features $X_j$, $j=1..M$, and a subset of labels $Y_i
\subseteq L$, where $L=\{y_1, y_2, \ldots y_q\}$ is the set of $q$
labels. Table~\ref{tab:multilabel} shows this representation. In
this scenario, the multi-label classification task consists of
generating a classifier $H$, which given an unseen instance $E =
(\mx, ?)$, is capable of accurately predicting its subset of labels
$Y$, \ie, $H(E) \rightarrow Y$.

\begin{table}[!htbp]
\renewcommand{\arraystretch}{1.3}
\caption{Multi-label data} \label{tab:multilabel} \centering
\scriptsize
 \begin{tabular}{c| c c c c c c |c}
  \cline{2-8}
             & &$X_1$    & $X_2$    & \ldots  & $X_M$    & & $Y$ \\ 
\hline \hline
$E_1$      &  & $x_{11}$ & $x_{12}$ & \ldots  & $x_{1M}$ & & $Y_1$  \\
$E_2$      &  & $x_{21}$ & $x_{22}$ & \ldots  & $x_{2M}$ & &$Y_2$  \\
\vdots     &  & $\vdots$ & $\vdots$ & $\ddots$& $\vdots$ & & $\vdots$ \\
$E_N$      &  & $x_{N1}$ & $x_{N2}$ & \ldots  & $x_{NM}$ & & $Y_N$
\\ \hline \hline
\end{tabular}
\end{table}

The predominant approaches of multi-label learning methods are:
algorithm adaptation and problem
transformation~\citep{Tsoumakas2010DataMining}. The first one
consists of methods which extend specific learning algorithms in
order to handle multi-label data directly.  The second approach is
algorithm independent, allowing the use of any state-of-the-art
single-label learning algorithm to carry out multi-label learning.
It consists of methods which transform the multi-label
classification problem into either several binary classification
problems, such as the \textit{Binary Relevance} ($BR$) approach, or
one multi-class classification problem, such as the \textit{Label
Powerset} ($LP$) approach.

The $BR$ approach decomposes the multi-label learning task into $q$
independent binary classification problems, one for each label in
$L$. To this end, the multi-label dataset $D$ is first decomposed
into $q$ binary datasets $D_{y_j}$, $j = 1..q$ which are used to
construct $q$ independent binary classifiers. In each binary
classification problem, examples associated with the corresponding
label are regarded as positive and the other examples as negative.
Finally, to classify a new multi-label instance $BR$ outputs the
aggregation of the labels positively predicted by the $q$
independent binary classifiers.  As $BR$ scales linearly with size
$q$ of the label set $L$, it is appropriate for not  a very large
$q$. Although in its simple form it experiences the deficiency in
which correlation among the labels is not taken into account,
successful  attempts have been made to model correlation using
binary
classifiers~\citep{Read09,Tsoumakas2009b,ChermanMM12}.
On the other hand, the $LP$ approach transforms the multi-label
learning task into a
 multi-class learning task considering every
 unique combination of labels  in a multi-label dataset as one
 class value of the corresponding multi-class dataset.
 Unlike $BR$,  $LP$ takes into account correlation among the labels.

Evaluating the performance of multi-label classifiers is difficult
mostly because multi-label prediction has an additional notion of
being \textit{partially} correct. To this end, several  measures  have been
proposed for the evaluation of bipartitions and rankings with
respect to the ground truth of multi-label data. A complete
discussion on these performance measures  is out of the scope of 
this paper, and can be found in \citep{Tsoumakas2010DataMining}. 

Measures that evaluate bipartitions are further divided into
\textit{example-based} and \textit{label-based}. The former are
calculated based on the average differences of the classifier
predicted multi-label of all examples in the test set, while the
latter decompose the evaluation process into separate evaluations of
each of the $q$ labels, which are afterwards averaged on all
labels. In what follows, we briefly describe the  measures used in this work to
evaluate bipartitions.

\subsection{Example-based}

Let $Y_i$ be the set of true labels (true multi-label) and $Z_i$ be the set of
predicted labels (predicted multi-label). \textit{Hamming-Loss} is defined by
Equation~\ref{eq:hammingloss}, where $\Delta$ represents the
symmetric difference between two sets.

\begin{equation}
    Hamming\textnormal{-}Loss(H,D)=\frac{1}{N}\sum^{N}_{i=1}{\frac{|Y_i\Delta Z_i|}{|L|}}
    \label{eq:hammingloss}
\end{equation}

\noindent \textit{Hamming-Loss} evaluates the frequency that labels
in the multi-label are  misclassified, \ie,  the  example  is
associated  to a  wrong label  or  a  label  belonging  to  the true
instance which is not predicted.

 \textit{Subset-Accuracy} is defined by
 Equation~\ref{eq:subsetaccuracy}, where $I$(true) = 1 and $I$(false) = 0.

\begin{equation}
    Subset\textnormal{-}Accuracy(H,D)=\frac{1}{N}\sum^{N}_{i=1}{I(Z_i = Y_i)}
    \label{eq:subsetaccuracy}
\end{equation}

\noindent
\textit{Subset-Accuracy} is a very strict evaluation
measure as it requires an exact match of the predicted and the true
set of labels.

In \citep{GodboleS04}, the following definitions for
\textit{Accuracy}, \textit{Precision} and \textit{Recall}, defined
by Equations~\ref{eq:accuracy}, \ref{eq:precision} and
\ref{eq:recall} respectively,   are proposed.

\begin{equation}
Accuracy(H,D)=\frac{1}{N}\sum^{N}_{i=1}{\frac{|Y_i \cap Z_i|}{|Y_i\cup Z_i|}} \label{eq:accuracy}
\end{equation}

\begin{equation}
  Precision(H,D)=\frac{1}{N}\sum^{N}_{i=1}{\frac{|Y_i \cap Z_i|}{|Z_i|}}
\label{eq:precision}
\end{equation}

\begin{equation}
  Recall(H,D)=\frac{1}{N}\sum^{N}_{i=1}{\frac{|Y_i \cap Z_i|}{|Y_i|}}
\label{eq:recall}
\end{equation}

\textit{Accuracy} is the proportion of the correctly predicted labels
to the total number of labels in the predicted and the truth label
set of an instance. \textit{Precision} is the proportion of
correctly predicted labels to the total number of predicted labels,
 and \textit{Recall} is the proportion
of correctly predicted labels to the total number of true labels.

\textit{F-Measure}, frequently used as performance measure for
information retrieval systems,  is the harmonic mean of
\textit{Precision} and \textit{Recall},  defined by
Equation~\ref{eq:f1}.

\begin{equation}
   F\textnormal{-}Measure(H,D)=\frac{1}{N}\sum^{N}_{i=1}{\frac{2 \times |Y_i \cap Z_i|}{|Y_i| + |Z_i|}} \label{eq:f1}
\end{equation}

All these performance measures have values in the interval $[0..1]$.
For \textit{Hamming-Loss}, the smaller the value, the better the
multi-label classifier performance is, while for the other measures,
greater values indicate better performance.

\subsection{Label-based}

In this case,  for each single label $y_i \in L$,   the $q$ binary
classifiers are initially  evaluated using any one of the binary
evaluation measures proposed in the literature, such as Accuracy,
F-Measure, ROC and others, which are afterwards averaged over all
labels. Two operations, \textit{macro-averaging} and
\textit{micro-averaging}, can be used to average over all labels.

Let
$B\left({T_{P_{y_i}},F_{P_{y_i}},T_{N_{y_i}},F_{N_{y_i}}}\right)$ be
a binary evaluation measure calculated for a  label $y_i$ based on
the number of true positive ($T_P$), false positive ($F_P$), true
negative ($T_N$) and false negative ($F_N$). The macro-average
version of $B$ is defined by Equation~\ref{eq:macro} and the
micro-average by Equation~\ref{eq:micro}.

\begin{equation} \label{eq:macro}
    B_{macro} = \frac{1}{q}\sum_{i=1}^{q}B\left({T_{P_{y_i}},F_{P_{y_i}},T_{N_{y_i}},F_{N_{y_i}}}\right)
\end{equation}

\begin{equation} \label{eq:micro}
 B_{micro} = B\left(  \sum_{i=1}^{q}{T_{P_{y_i}}},\sum_{i=1}^{q}{F_{P_{y_i}}},\sum_{i=1}^{q}{T_{N_{y_i}}},\sum_{i=1}^{q}{F_{N_{y_i}}} \right)
\end{equation}

Thus,   the binary evaluation measure used is computed on individual
labels first and then averaged for all labels by the
macro-averaging operation, while  it is computed globally for all
instances and all labels by the micro-averaging operation. This
means that macro-averaging would be more affected by labels that
participate in fewer multi-labels, \ie, fewer examples, which is
appropriate in the study of unbalanced
datasets~\citep{DendamrongvitVK11}.  Furthermore,
it should be observed that for some binary evaluation measures, such
as Accuracy, macro-average and micro-average yield the same result.

\section{The Simple Multi-label Baseline Classifier \lbg} \label{sec:lgb}

In supervised learning, simple baseline classifiers can be constructed by 
only looking at the class, \ie, ignoring any other information from the 
dataset.  The single-label learning community frequently uses as a reference 
the one which always predicts the majority class (the most frequent class value). 
Although a classifier might perform worse than this simple baseline classifier, as could be the case when learning from highly skewed domains, this behaviour requires a special explanation.
In multi-label learning, as the \lp transformation maps each distinct
multi-label into a single-label, transforming the multi-label
dataset into a multi-class (single-label) dataset, one could argue
\textit{why not use the one which always predicts the majority multi-label as the most simple multi-label baseline classifier?}.
Although it is a possible baseline, which focuses on maximizing
 \subacc, this strategy does not take into account  the individual
label distribution in  the multi-labels, which provides additional
information.

Moreover, due to the fact that multi-label prediction has the
notion of being partially correct, several multi-label
evaluation measures have been proposed to evaluate the
classification performance from different perspectives.

 In~\citep{MetzACM12}, we propose
specific simple baseline classifiers which are tailored to
maximize/minimize one specific multi-label measure at a time.
However, a specific baseline classifier tailored to
maximize/minimize one measure does not necessarily maximize/minimize
the other measures. Nevertheless, having
different baseline classifiers to consider would be  a cumbersome
task, due to the number of different
multi-label evaluation measures proposed in the literature, as well
as multi-label learning algorithms which are tailored to
maximize/minimize more than one measure (multi-objective). In this work we also
proposed \lbg, a simple baseline classifier which does not focus in
maximizing/minimizing any specific measure,  and which can be used
to find general baselines for any bipartite multi-label evaluation
measure.

The rationale behind  \lbg to find the predicted multi-label $Z$ is
very simple. It consists of ranking the single-labels
in $L$ according to their individual relative frequencies in the
multi-labels, and  then, the $\sigma$ most frequent single-labels are
included in $Z$. We are then left with the problem of choosing
$\sigma$ such that $Z$ is representative, \ie, with a reasonable
number of single-labels and at the same time avoiding being too
strict (including too few single-labels) or too flexible (including
too many single-labels). As we are interested in finding $Z$ that
 represents the single-label distribution in the  multi-labels well,
 we defined $\sigma$ as the closest integer value of the label cardinality.   The label
cardinality, defined by Equation~\ref{Eq:card}, represents the
average size of the multi-labels in a dataset $D$ composed of $N$ examples, \ie, the average
number of single-labels associated to each instance.

\begin{equation}
    CR(D) = \frac{1}{N}\sum^{N}_{i=1}{|Y_i|}
    \label{Eq:card}
\end{equation}

In case of ties (single-labels with the same frequency), we consider
the label co-\-oc\-cur\-rence measure, choosing the one which maximizes
its co-\-oc\-cur\-rence with  the better ranked labels. It should be observed that, as every other learner, \lbg has a particular bias. For instance, it will work well for \subacc whenever there is a positive correlation among the most frequent labels. However, it will not work well when the correlation is negative. In other words, it will work better whenever its bias fits the dataset well.

The specific baseline classifiers, as well as  \lbg, were
implemented using the Mulan framework~\citep{tsoumakas2011mulan}, a
Java package for multi-label classification based on
Weka\footnote{{\url{http://www.cs.waitako.ac.nz/ml/weka}}}, which is
commonly used by the multi-label learning community.

 An analysis of several multi-label bipartition evaluation measure
 baselines obtained by the specifics, as well as by the \lbg baseline classifier showed that, as expected, the specific ones perform better on the
measure they try to maximize/minimize, although they degrade on the
other measures. On the other hand, \lbg shows a reasonable performance for
all the considered bipartite measures. Ranking the results obtained
by the specific baseline classifiers and by \lbg, it was observed
that \lbg is ranked ``in the middle'', as shown in
~\citep{MetzACM12}, making it suitable to be used
as a general baseline multi-label classifier.


\section{Systematic Review}
\label{sec:systematicreview}
 Although multi-label classification has
drawn increasing attention from the machine learning and data
mining communities in the past decade, there are few extensive
reviews researching publications on this topic. Moreover, to the best
of our knowledge, there is no extensive review which explicitly
focused on papers reporting  experimental results for multi-label
learning.

To this end, as we need published experimental results on evaluation
measures of multi-label classifiers to compare with our proposed
multi-label baseline classifier \lbg, we have carried out the
Systematic Review (SR) process, a method to search for relevant
papers in a wide, rigorous and replicable
way~\citep{KitchenhamPBBTNL10}. The SR process
is able to answer Research Questions (RQ) about a subject by using a
protocol of planned activities to identify, select and summarize
relevant pieces of work.

The aim of our systematic review, which is reported
in~\citep{spolaor2013systematic}, is to answer the following RQ:
\textit{what are the publications which report experimental results
for multi-label learning research?}. To this end, we used nine world wide online bibliographic database
search engines as listed in Table~\ref{tab:databasessearched}, in which 1,543 publications were
identified.

\begin{table}[hbt]
\centering
  \caption{Bibliographic databases used in the systematic review}
  \scriptsize
    \begin{tabular}{ll}
     \addlinespace
    \toprule
    Database & URL \\
        \midrule
ACM Portal  & \url{http://portal.acm.org} \\
CiteSeer    & \url{http://citeseerx.ist.psu.edu}\\
Interscience  & \url{http://onlinelibrary.wiley.com} \\
ScienceDirect & \url{http://www.sciencedirect.com} \\
Scirus        & \url{http://scirus.com}  \\
Scopus        & \url{http://www.scopus.com} \\
SpringerLink  & \url{http://link.springer.com} \\
Xplore      & \url{http://ieeexplore.ieee.org} \\
Web of Science & \url{http://isiknowledge.com} \\
    \bottomrule
    \end{tabular}
  \label{tab:databasessearched}
\end{table}

Some retrieved pieces of work can be duplicated,  as some sources,
\ie, journals, proceedings and others, are indexed by more than one
bibliographic database. Thus, cases with duplicated titles were
automatically or manually (mistyped titles) removed, keeping only
one copy of the publication. From the 1,543 publications, 847 (55\%)
were automatically removed and 79 (5\%) were manually removed. Thus,
we were left  with 617 (40\%) publications which were divided among
the authors of this paper, such that each one of the 617
publications was manually analyzed using 16 exclusion
criteria as a guide. Whenever a  publication fulfilled one or more exclusion criteria, it was
removed. If there were doubts about removing a publication, a second
reviewer verified the doubtful publication.

\begin{figure}[!htb]
\centering 
\includegraphics[width=0.9\textwidth]{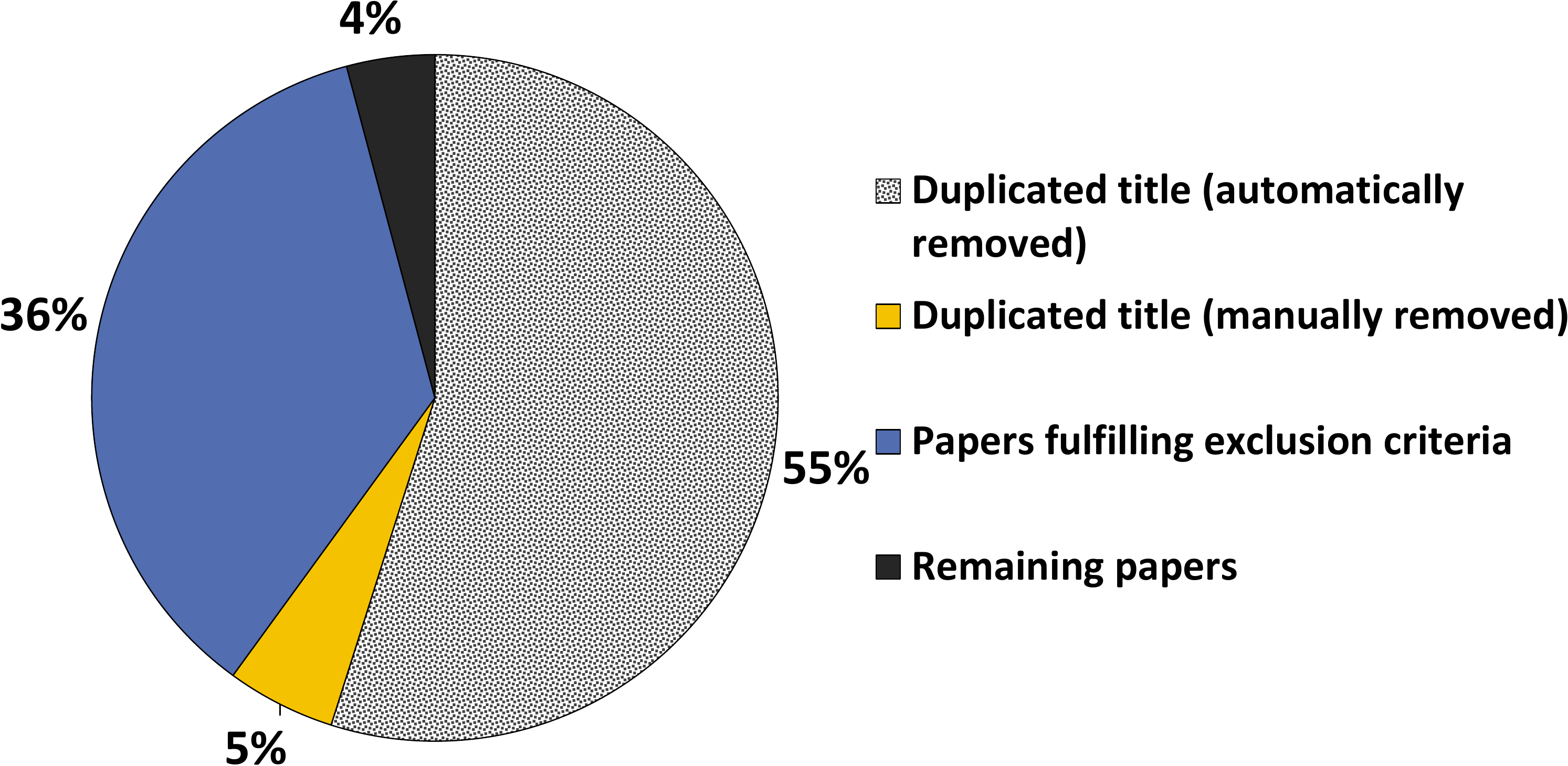}
\caption{Summary of the selection procedures to find the relevant
papers used in this work from a total of 1,543 publications}
\label{fig:paperselection}
\end{figure}

The 16 exclusion criteria include: publications which do not
consider example-based or label-based evaluation measures;
restricted access to the dataset; pre-processed datasets where the
final attribute-value table used by the learning algorithm is not
publicly available, and others. Recall that we only collected evaluation
measures of classifiers that were obtained by multi-label learning
algorithms using identical attribute-value datasets. At this
stage, we were left with 64 (4\%) publications which do not fulfill
any of the 16 exclusion criteria. Figure~\ref{fig:paperselection} shows a
summary of the selection procedures.

Nevertheless, similar to known systematic
reviews~\citep{KitchenhamPBBTNL10}, results are
bound to the electronic databases searched for publications, nine
in our case. Thus, papers potentially relevant to the research
question might  not have been identified.

From these 64 papers, we recorded information extracted manually on
 an electronic spreadsheet with 42 columns,
described in detail in~\citep{spolaor2013systematic}. As most of
the information extraction has to be carried out manually, this
process was double checked. A relational database was set up to
appropriately record the 42 columns, modelling each sheet as a
database table. In this database, each sheet column is a table
attribute and each sheet line is an instance.  The corresponding
entity-relationship model consists of four tables: \textit{main},
\textit{dataset}, \textit{measure} and  \textit{paper},  as well  as
some relationships between them. The \textit{main} table records the
experimental settings and results published in the papers which are
able to answer the research question, as well as some foreign keys which
link results to a paper and a dataset. Furthermore, the
\textit{dataset} table records usual statistics from multi-label
datasets, such as: the domain; number of examples; features and
labels; as well as the number of different multi-labels, label
cardinality and label density; the URL where the dataset is publicly
available is also kept in this table. The \textit{measure} table
manages the name and type of the recorded multi-label evaluation
measures. The \textit{paper} table records the selected
publications.

\begin{figure}[!htb]
\centering
\includegraphics[width=0.65\textwidth]{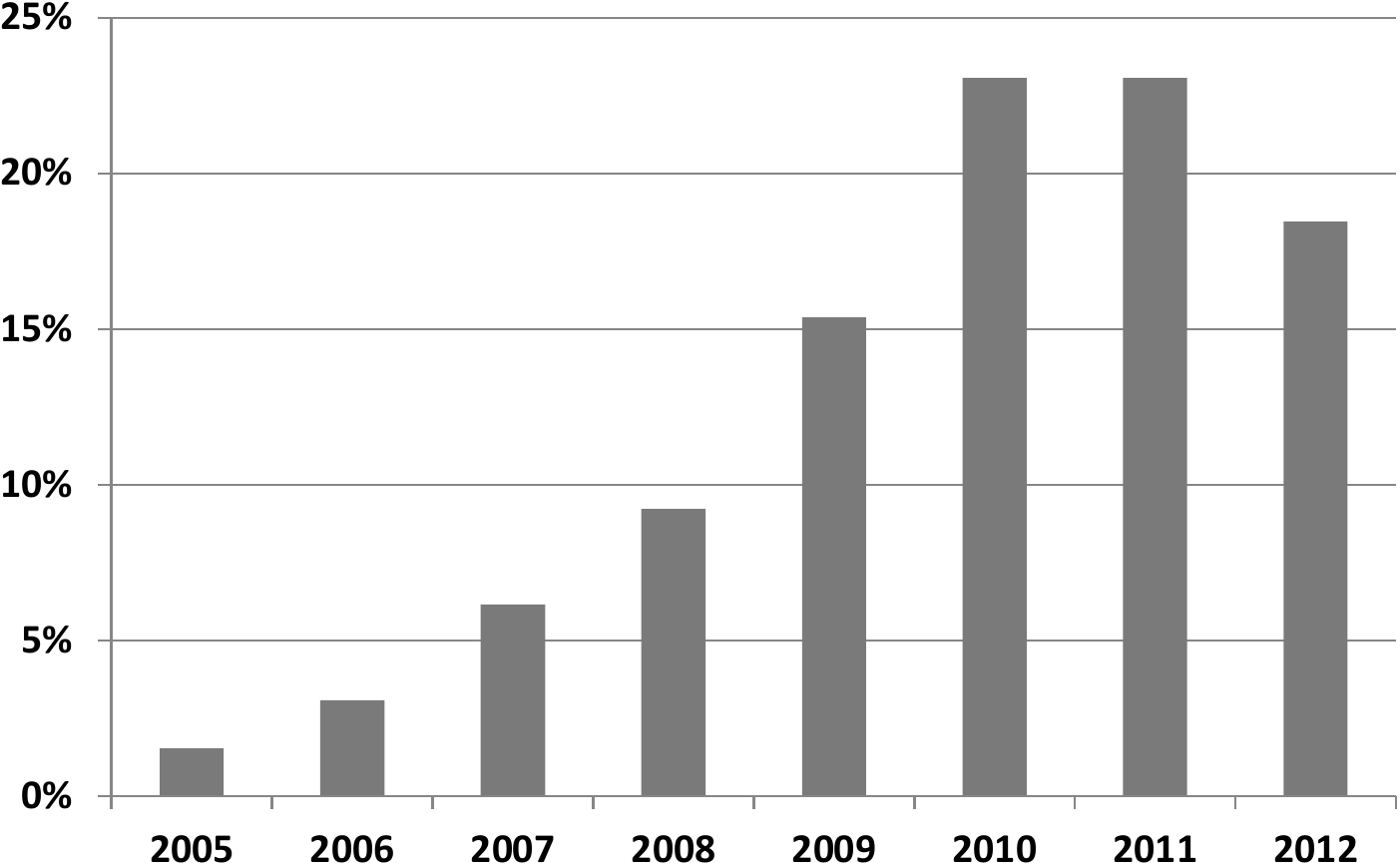}
\caption{Percentage of the 64 papers published per year}
\label{fig:paperdistbyyear}
\end{figure}

Figure~\ref{fig:paperdistbyyear} shows the distribution per year of the 64
papers considered in this work, where 21 (33\%) of them were
published in journals and the remaining in congresses and workshops.
Moreover, besides 7 (10\%) papers published in the Machine Learning
Journal, at most 2 papers  were  published  in the same source.

\begin{figure}[!htb]
\centering
\includegraphics[width=1\textwidth]{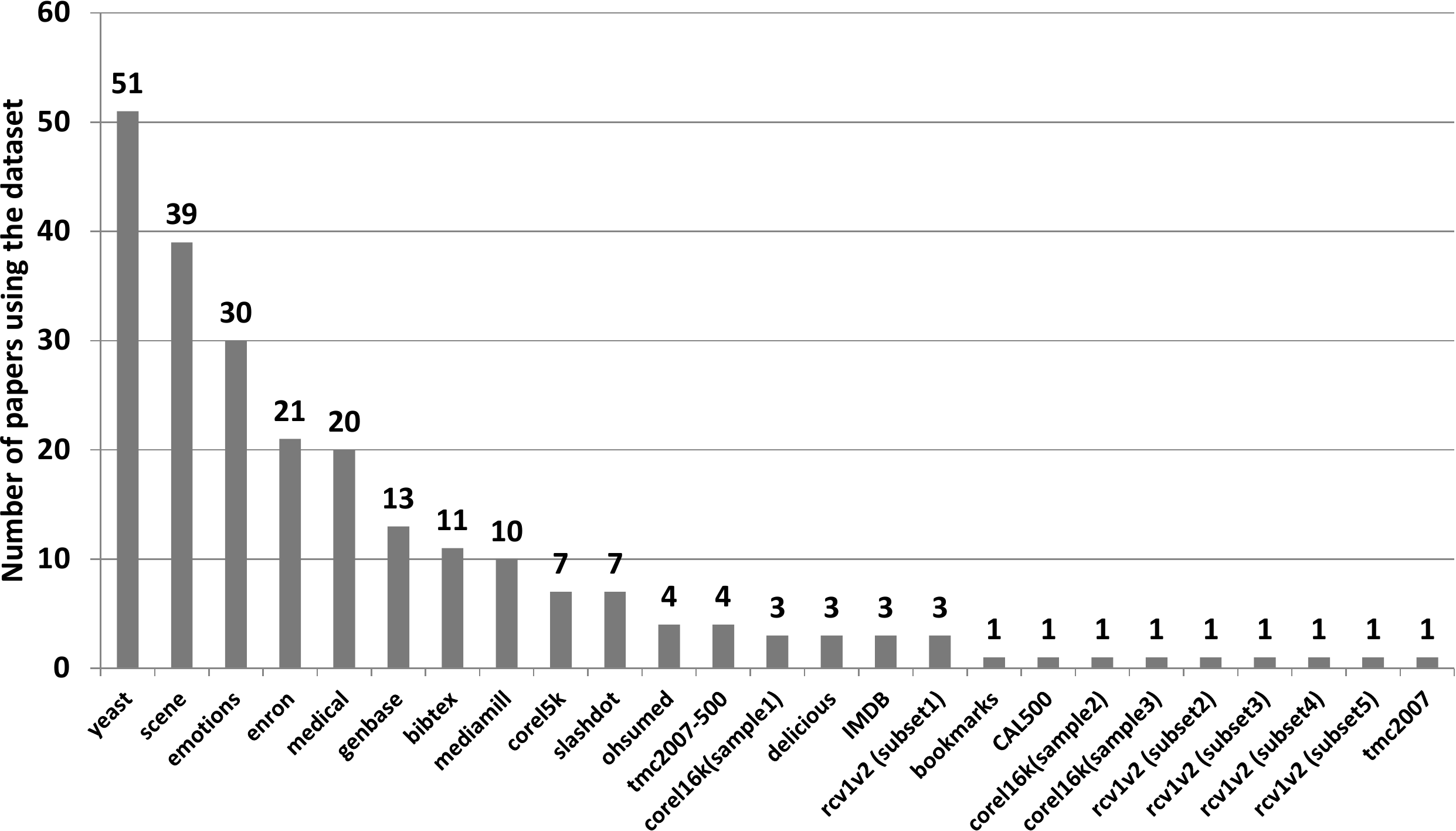}
\caption{Number of papers using each multi-label dataset}
\label{fig:papersPerDatasets}
\end{figure}

Figure~\ref{fig:papersPerDatasets} shows the number of papers in which each
dataset was used.  As already mentioned, we do not consider
experimental results from pre-processed datasets, such as the very
frequently used Reuters, unless the final attribute-value table from
which the classifier is generated is reported. Thus, few  datasets
whose original domain is
 text were considered in this study due to this
restriction.  As can be observed in this figure, the \yeast dataset
is used in almost 80\% of the 64 papers considered in this work.

\section{Comparing \lbg to Published Evaluation Measure Values} \label{sec:expdesign}

In this section, some statistics of published experimental
evaluation measure values of multi-label classifiers and the ones
obtained by \lbg are discussed.

\subsection{Datasets} \label{sec:datasets}

From the 25 datasets used in the 64 papers selected by the
systematic review process  shown in Figure~\ref{fig:papersPerDatasets},
 the 10 most frequently used are the ones
considered.

Table~\ref{tab:dataset} presents the selected datasets and associated statistics. It shows: the application domain (Domain);  number of instances (\#E);  number  of features (\#F); the  total number of labels ($|L|$);  and the percentage of the 64 publications which use  the  dataset (Usage). Moreover, Table~\ref{tab:labels} shows some statistics associated with the datasets labels: label cardinality ($CR(D)$), defined by Equation~\ref{Eq:card};  label density ($DS(D)$), defined by Equation~\ref{Eq:dens};   number of distinct multi-labels (\#Dist); the lowest (Min) and the highest (Max) single-label frequencies, as well as the first (1Q), second (median Med) and  the third quartiles (3Q), as suggested by \cite{TsoumakasPersonalCommunication2013}.

\begin{equation}
    DS(D) = \frac{1}{N}\sum^{N}_{i=1}{\frac{|Y_i|}{|L|}}
    \label{Eq:dens}
\end{equation}

%

\begin{table}[htbp]
 \centering
 \caption{Datasets and associated statistics}
 \scriptsize
 \begin{tabular}{lccccc}
 \addlinespace
 \toprule
 Dataset		& Domain	& \#E	& \#F	& $|L|$	& Usage \\
 \midrule
\bibtex$^1$		& text		& 7395	& 1836 	& 159	& 17\%  \\
\corel$^1$		& image		& 5000	& 499  	& 374	& 11\%  \\
\emotions$^1$	& music  	& 593	& 72 	& 6 	& 47\%  \\
\enron$^1$		& text 		& 1702	& 1001 	& 53 	& 33\% \\
\genbase$^1$	& biology	& 662	& 1186	& 27 	& 20\% \\
\mediamill$^1$	& video		& 43907	& 120 	& 101	& 16\% \\
\medical$^1$	& text 		& 978	& 1449 	& 45 	& 31\% \\
\scene$^1$		& image		& 2407	& 294  	& 6 	& 61\% \\
\slashdot$^2$	& text		& 3782	& 1079 	& 22 	& 11\% \\
\yeast$^1$		& biology	& 2417	& 103  	& 14 	& 80\% \\
\midrule
web link:	& \multicolumn{5}{l}{$^1$\url{http://mulan.sourceforge.net/datasets.html}} \\
	& \multicolumn{5}{l}{$^2$\url{http://meka.sourceforge.net/} } \\
 \bottomrule
 \end{tabular}
 \label{tab:dataset}
\end{table}

\begin{table}[htbp]
 \centering
 \caption{Labels' associated statistics}
 \scriptsize
 \begin{tabular}{lcccccccc}
 \addlinespace
 \toprule
 Dataset		& $CR(D)$	& $DS(D)$	& \#Dist	& Min	& 1Q	& Med	& 3Q	& Max	\\
 \midrule         
\bibtex		& 2.402		& 0.015	& 2856	& 51	& 61	& 82	& 129	& 1042			\\
\corel		& 3.522		& 0.009	& 3175	& 1		& 6	& 15	& 39	& 1120				\\
\emotions	& 1.869		& 0.311	& 27	& 148	& 166	& 170	& 185	& 264			\\
\enron		& 3.378		& 0.064	& 753	& 1		& 13	& 26	& 107	& 913			\\
\genbase	& 1.252		& 0.046	& 32	& 1		& 3	& 17	& 49	& 171				\\
\mediamill	& 4.376		& 0.043	& 6555	& 31	& 93	& 312	& 1263	& 33869			\\
\medical	& 1.245		& 0.028	& 94	& 1		& 2	& 8	& 34	& 266					\\
\scene		& 1.074		& 0.179	& 15	& 364	& 404	& 429	& 432	& 533			\\
\slashdot	& 1.181		& 0.054	& 156	& 0		& 26	& 179	& 250	& 584			\\
\yeast		& 4.237		& 0.303	& 198	& 34	& 324	& 659	& 953	& 1816			\\

 \bottomrule
 \end{tabular}
 \label{tab:labels}
\end{table}

Observe that these 10 datasets from five different domains have
different characteristics.  The number of instances vary from 593 up
to 43,907; the number of features from 72 up to 1,836 and the number
of single-labels ($|L|$) from 6 up to 374. Furthermore, the label
cardinality varies from 1.074 up to 4.376; the label density from
0.009 up to 0.311 and the number of distinct multi-labels from 15 up
to 6,555. It is worth noting that some datasets present labels with very low or zero frequency (Min). Although it could be a good practice to remove these sort of labels, the original versions of the datasets were kept in this work. More detail about these datasets can be found in the site where they are publicly available.

\subsection{Evaluation measures}

\begin{figure}[h]
\centering 
    \includegraphics[width=0.9\textwidth]{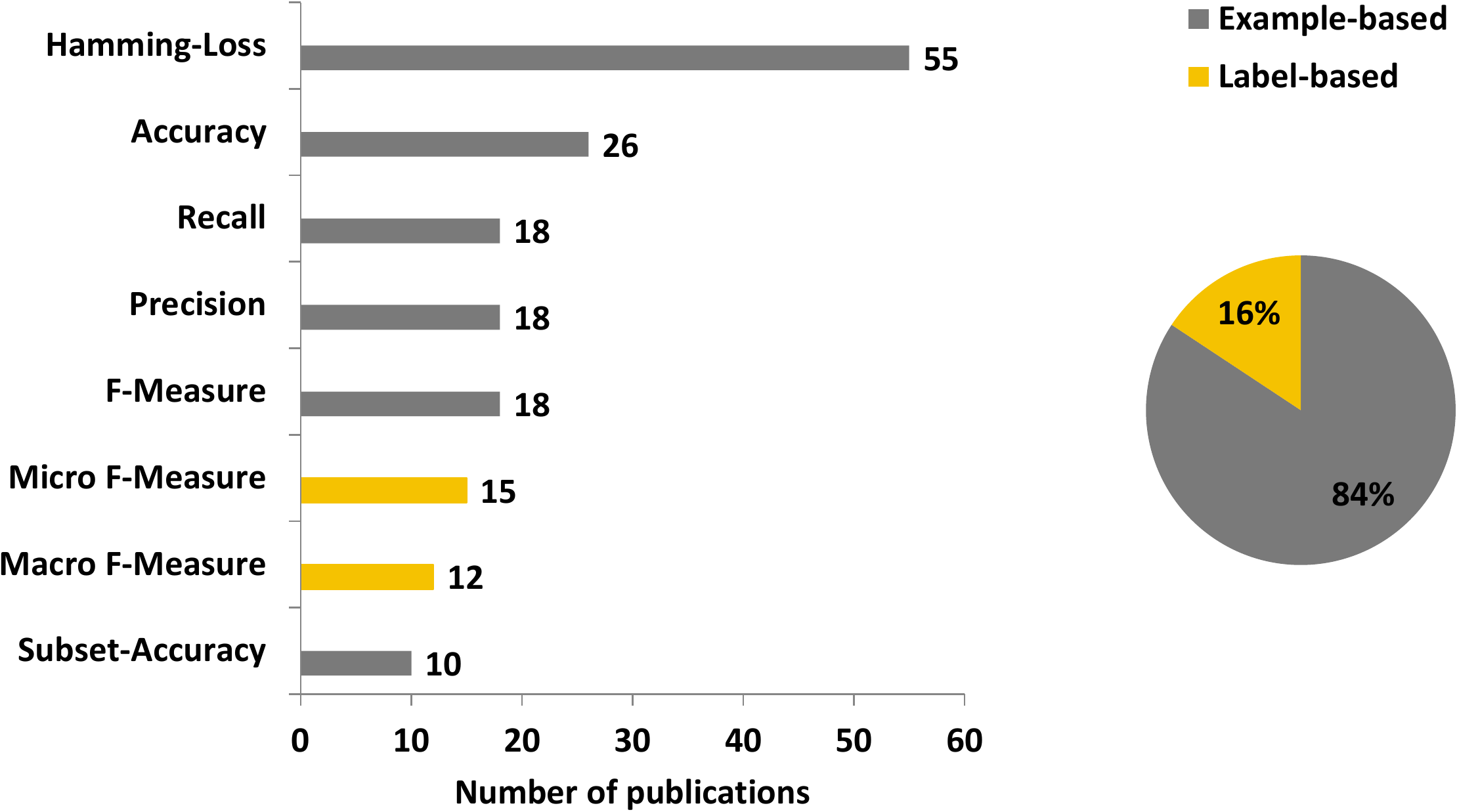}
\caption{Number of papers using each evaluation measure}
\label{fig:measuresUsage}
\end{figure}

As explained in Section~\ref{sec:systematicreview}, all
example-based and label-based measure values reported in the 64
papers were manually collected.  Similar  to the dataset selection,
we chose the 8 most frequently used out of the 17 different recorded
evaluation measures. Moreover, the 9 measures not considered here are used in very few (at most 5) of the
64 papers. Figure~\ref{fig:measuresUsage} shows the number of papers
in which the 8 evaluation measures considered were
used.

As can be observed, at least in the papers considered in this work,
example-based measures are much more frequently used than
the label-based ones. Furthermore, among the example-based measures, \hloss is
the most frequently used (55 papers), while \subacc is used in
fewer papers. As already mentioned, \hloss evaluates partially correct
classification, while \textit{Sub\-set-Ac\-cu\-ra\-cy}  evaluates exact matching between the
ground truth and the predicted multi-label.

The results reported on these measures come from different
experimental setup and validation processes, such as cross-validation
and hold-out. Considering all the 64 papers, it was observed that
$49.6\%$ were obtained using hold-out validation, $43.0\%$ using
some type of $k$-fold cross-validation, such as  10-fold, 3-fold or
$5\times2$-fold,  and the rest using other validation
process or the validation process is not explicitly mentioned in the
respective publication. However, it is interesting to note that from
\textit{all} the measure values which we have manually extracted and
recorded, a total of 6,989, the standard deviation is reported for
only 1,435 (20.5\%) of them.

Next, statistics related to the published measure values  considered
in this work, which were obtained from the classifiers generated by
a variety of multi-label learning algorithms, and the ones obtained
by \lbg, are presented.

\subsection{Results and discussion}
\label{sec:results}

Table~\ref{tab:baselines} shows the  \lbg baseline values for each evaluation measure and dataset considered in this work. The eight measures are denoted as: \textit{Example-based Accuracy} (\acc);   \textit{Example-based F-Measure} (\f); \textit{Example-based Hamming-Loss} (\hl); \textit{Example-based Precision} (\pr);    \textit{Example-based Recall} (\re); \textit{Example-based Subset-Ac\-cu\-ra\-cy} (\sacc); \textit{Label-based Macro-averaged F-Measure} (\fM);  and  \textit{Label-based Micro-averaged F-Measure} (\fm).

These values can be directly used in other publications, as they are the same for a given dataset and evaluation measure.

\begin{table}[h!]
  \centering
  \caption{\lbg baseline measure values}
  \scriptsize
    \begin{tabular}{lcccccccc}
    \addlinespace
    \toprule
Dataset & \acc & \f & \hl & \pr & \re & \sacc & \fM & \fm \\
    \midrule
\bibtex & 0.07 & 0.10 & 0.03 & 0.11 & 0.11 & 0.00 & 0.00 & 0.10 \\
\corel & 0.12 & 0.18 & 0.01 & 0.20 & 0.17 & 0.00 & 0.00 & 0.19 \\
\emotions & 0.23 & 0.30 & 0.33 & 0.45 & 0.23 & 0.07 & 0.10 & 0.31 \\
\enron & 0.30 & 0.42 & 0.07 & 0.48 & 0.39 & 0.00 & 0.04 & 0.45 \\
\genbase & 0.26 & 0.26 & 0.06 & 0.26 & 0.26 & 0.26 & 0.02 & 0.23 \\
\mediamill & 0.35 & 0.50 & 0.04 & 0.53 & 0.53 & 0.00 & 0.03 & 0.50 \\
\medical & 0.21 & 0.23 & 0.04 & 0.27 & 0.21 & 0.16 & 0.01 & 0.24 \\
\scene & 0.19 & 0.20 & 0.27 & 0.22 & 0.19 & 0.17 & 0.06 & 0.21 \\
\slashdot & 0.15 & 0.15 & 0.09 & 0.15 & 0.15 & 0.14 & 0.10 &  0.14 \\
\yeast & 0.42 & 0.55 & 0.26 & 0.58 & 0.55 & 0.05 & 0.21 & 0.57 \\
    \bottomrule
    \end{tabular}
  \label{tab:baselines}
\end{table}

\newcolumntype{g}{>{\columncolor{lightgray}}c}
\begin{table}[h!]
  \centering
  \caption{Number of measure values which underperform or are equal to the corresponding \lbg baseline value (sorted by \%)}
  \scriptsize

\resizebox{\textwidth}{!}{

    \begin{tabular}{lcccccccc  ggg}

    \addlinespace
    \toprule

Dataset & \acc & \f & \hl & \pr & \re & \sacc & \fM & \fm &  \#$U_d$ & \#$M_d$ & \%  \\
\midrule

\corel & 18 & 30 & 13 & 8 & 13 & 12 & 4 & 9 & 107 & 249 & 43.0 \\
\mediamill & 24 & 23 & 19 & 10 & 25 & 2 & 3 & 12 & 118 & 311 & 37.9 \\
\enron & 32 & 33 & 34 & 18 & 20 & 6 & 4 & 4 & 151 & 606 & 24.9 \\
\slashdot & 11 & 5 & 11 & 5 & 4 & 9 & 5 & 0 & 50 & 266 & 18.8 \\
\yeast & 17 & 20 & 52 & 27 & 23 & 2 & 3 & 11 & 155 & 1094 & 14.2 \\
\bibtex & 5 & 5 & 16 & 4 & 6 & 0 & 0 & 0 & 36 & 326 & 11.0 \\
\genbase & 6 & 3 & 4 & 5 & 3 & 0 & 0 & 0 & 21 & 346 & 6.1 \\
\medical & 6 & 4 & 5 & 6 & 2 & 3 & 0 & 0 & 26 & 540 & 4.8 \\
\scene & 3 & 1 & 4 & 2 & 3 & 1 & 0 & 0 & 14 & 888 & 1.6 \\
\emotions & 2 & 0 & 1 & 1 & 0 & 0 & 0 & 0 & 4 & 716 & 0.6 \\ 

\rowcolor{lightgray} \#$U_m$ & 124 & 124 & 159 & 86 & 99 & 35 & 19 & 36 &  &  & \\
\rowcolor{lightgray} \#$M_m$ & 907 & 782 & 1355 & 580 & 580 & 490 & 245 & 403 &  &  & \\
\rowcolor{lightgray} \% & 13.7 & 15.9  & 11.7  & 14.8  & 17.1  & 7.1  & 7.8 & 8.9  &  &  & \\

    \bottomrule
    \end{tabular}
    }
  \label{tab:globalperformance}
\end{table}

\begin{figure}[h!]
\centering
    \includegraphics[width=0.8\textwidth]{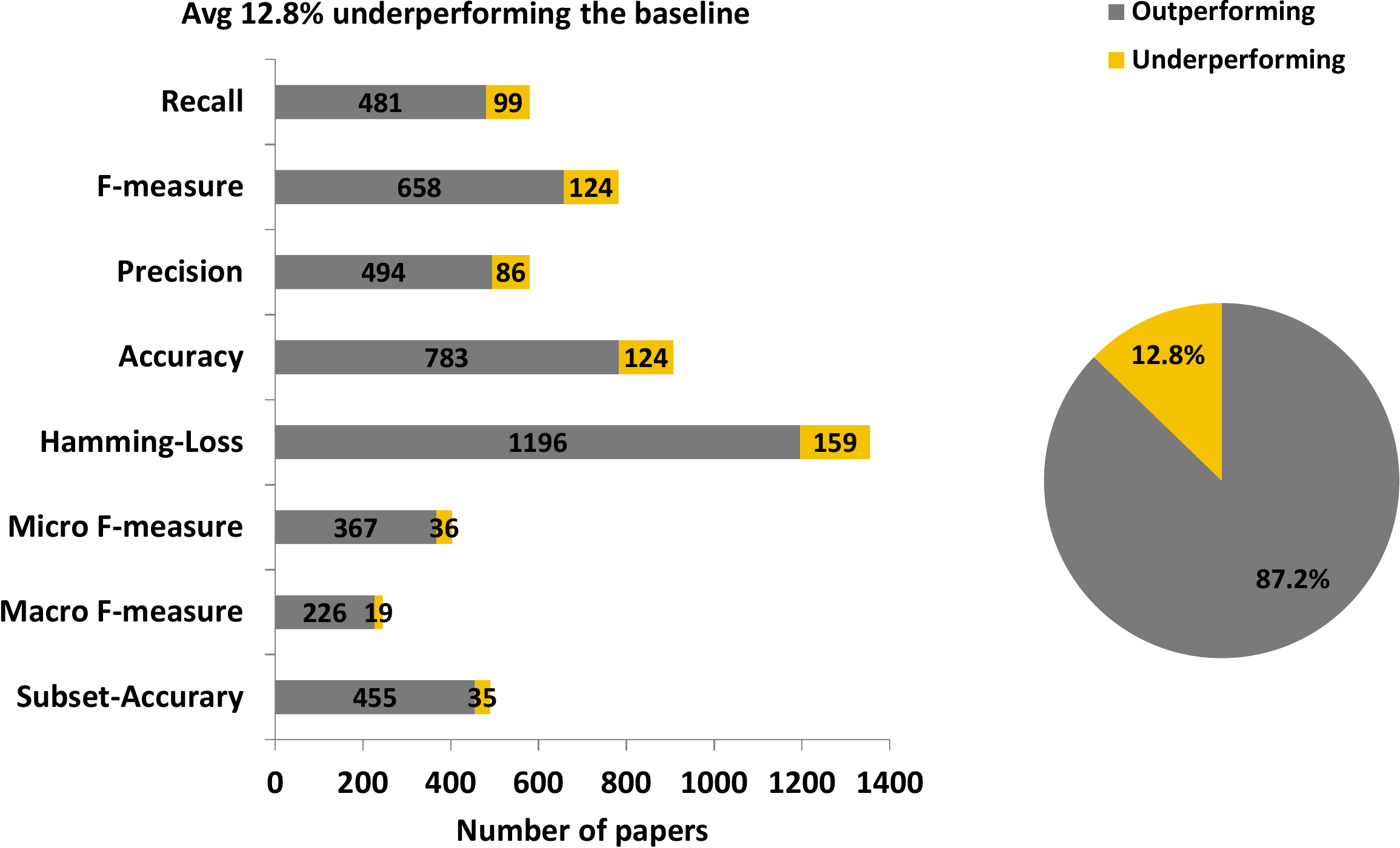}    
 \caption{Overall performance per measure using \lbg as reference}
\label{fig:overallPerformance}
\end{figure}

\begin{figure}[h!]
\centering 
    \includegraphics[width=0.47\textwidth]{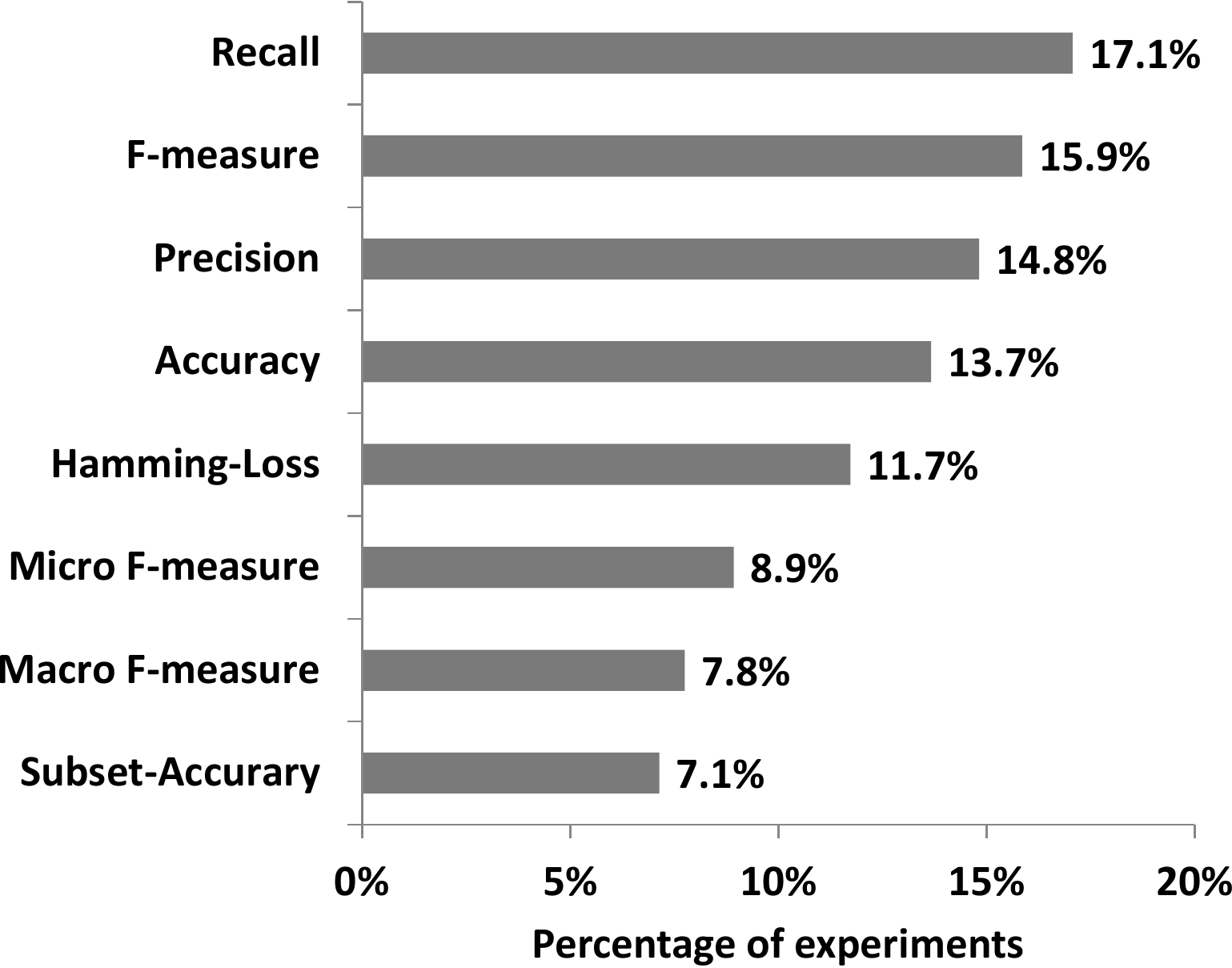}
    \includegraphics[width=0.47\textwidth]{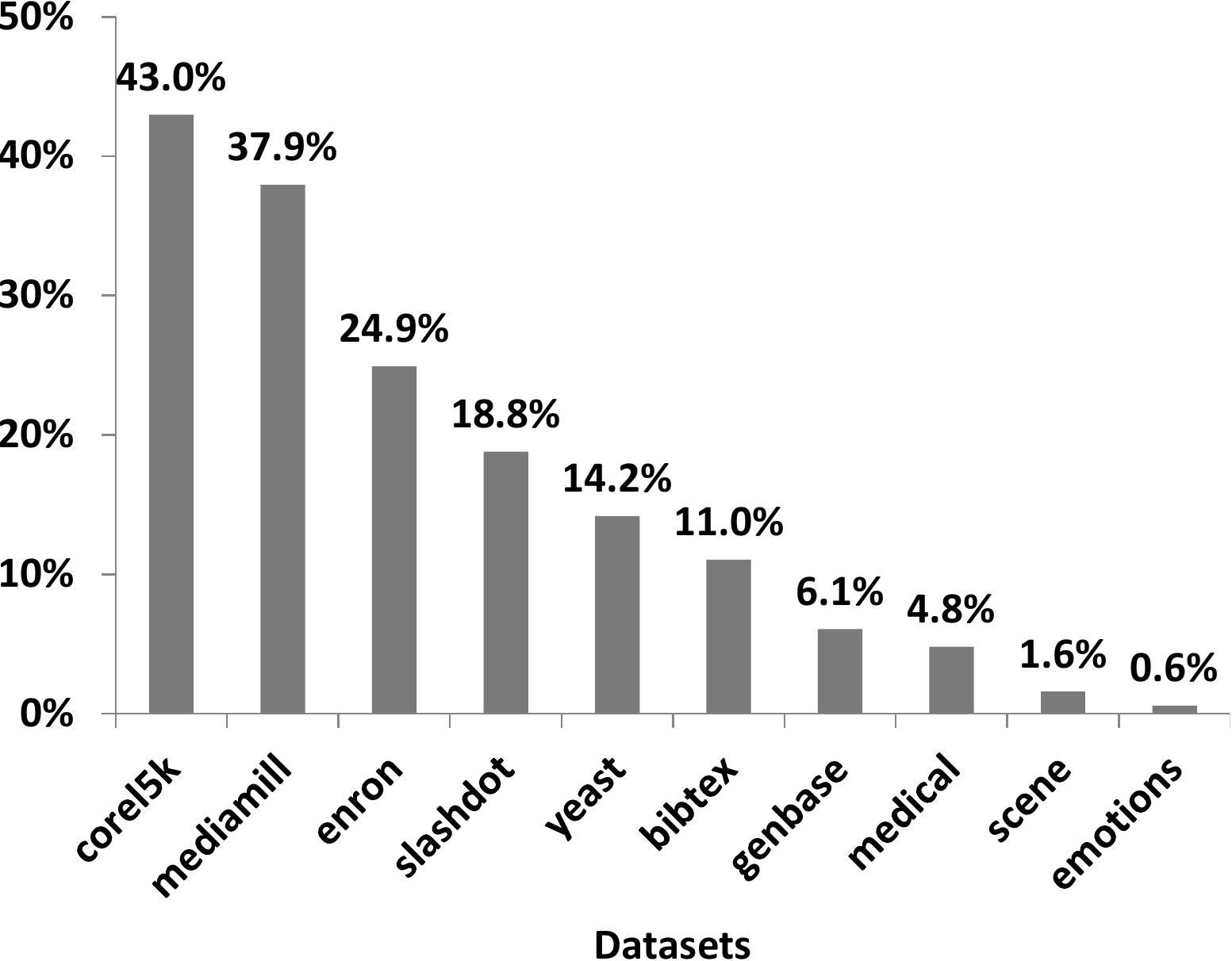}

\caption{Percentage of values per measure and
dataset which underperform or are equal to the corresponding \lbg
baseline value}
\label{fig:overallPerformancepercent}
\end{figure}

Table~\ref{tab:globalperformance} shows, for each dataset,   the
number of times that a published measure value underperforms or it
is equal to the corresponding \lbg baseline value. Summary
information is shown in light gray cells. Column \#$U_d$ shows the
total number of measures fulfilling this condition on a total of
\#$M_d$ measure values recorded for each dataset, and column \%
shows the percentage. Similar results are shown in rows \#$U_m$,
\#$M_m$ and \% for each measure considered. This information is shown graphically
in Figures~\ref{fig:overallPerformance} and~\ref{fig:overallPerformancepercent}.

\begin{figure}[ht]
\centering 
\subfigure[\corel]{
    \includegraphics[width=0.4\textwidth]{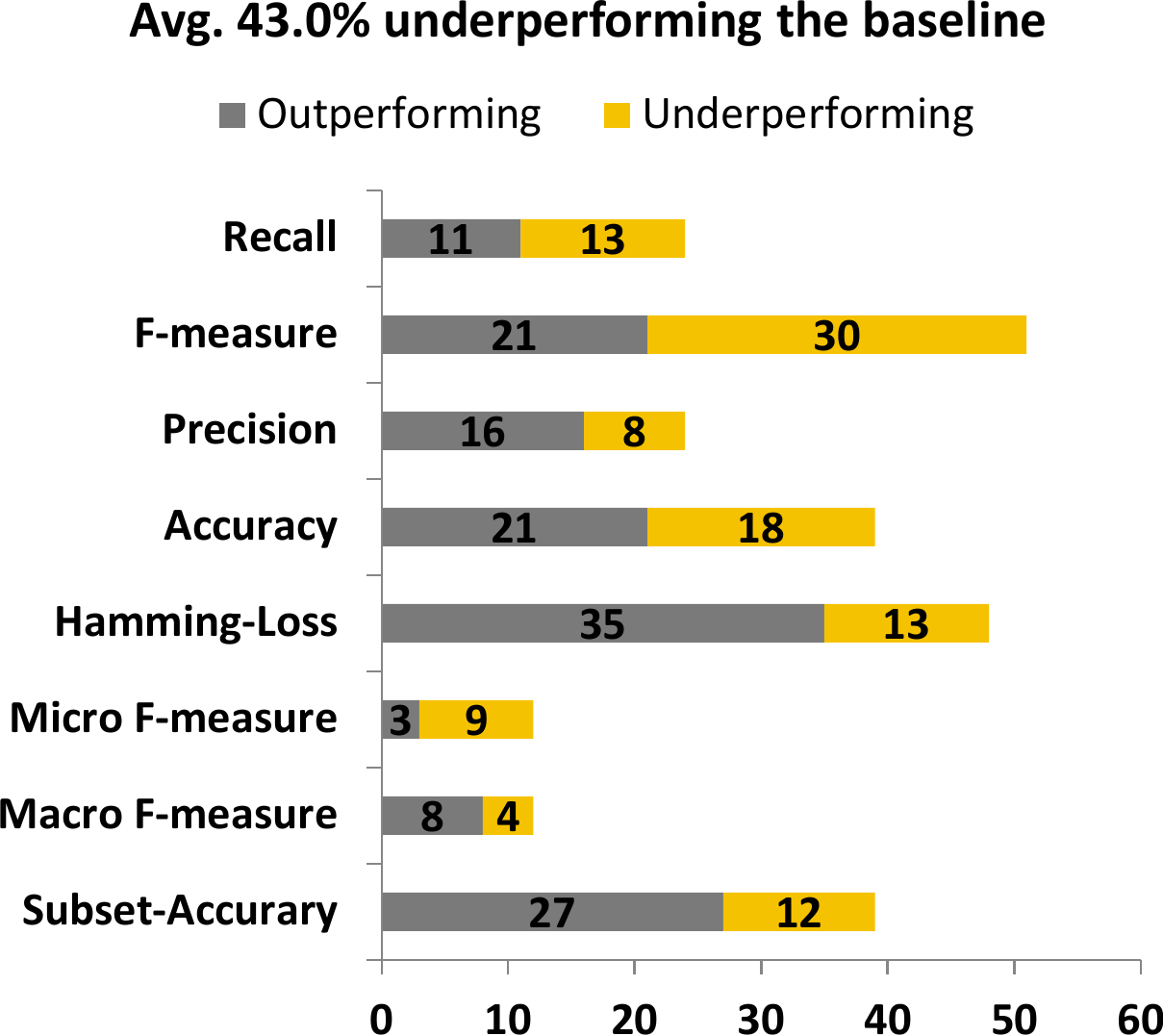}
} \subfigure[\mediamill]{
    \includegraphics[width=0.4\textwidth]{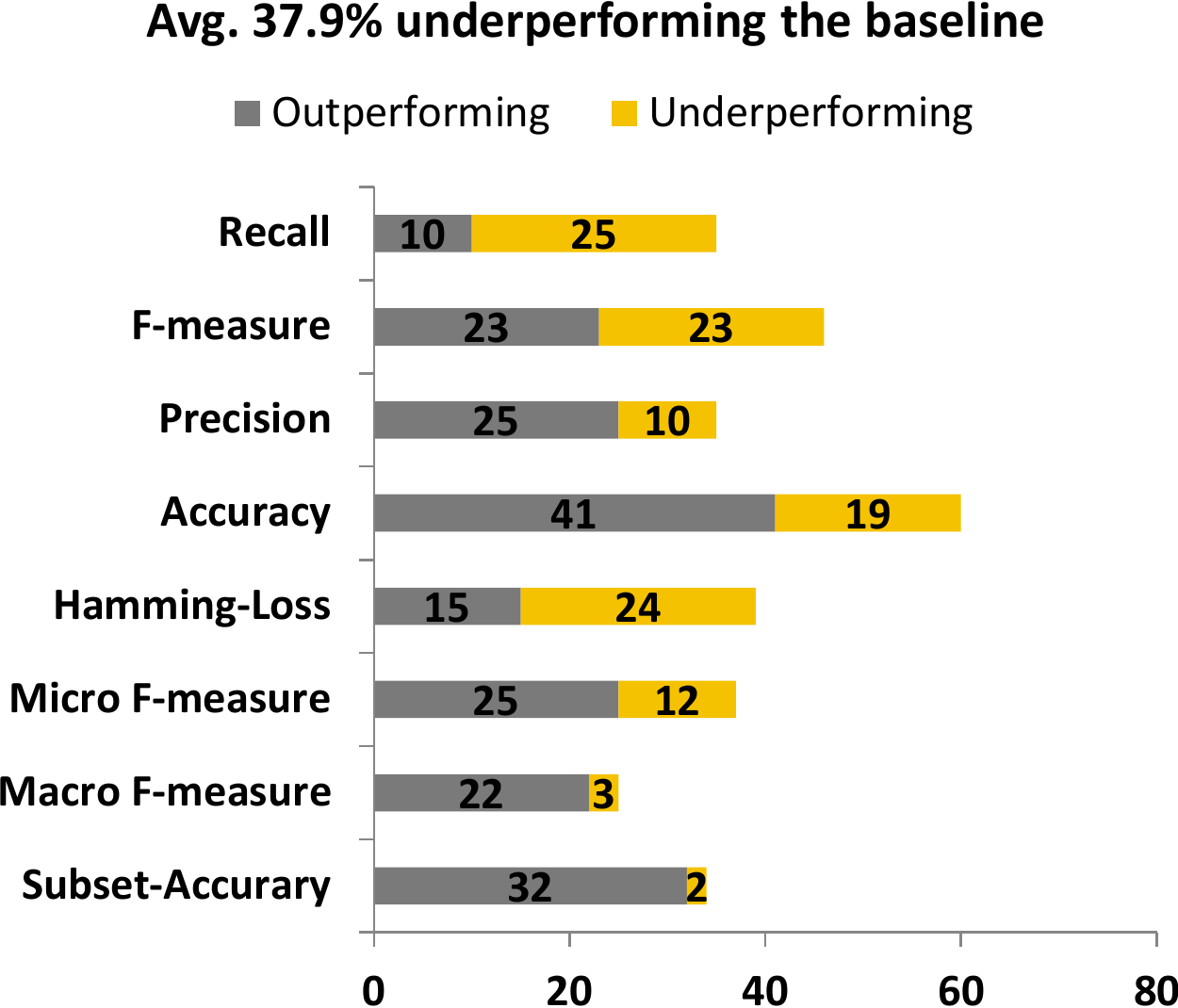}
} \subfigure[\enron]{
    \includegraphics[width=0.4\textwidth]{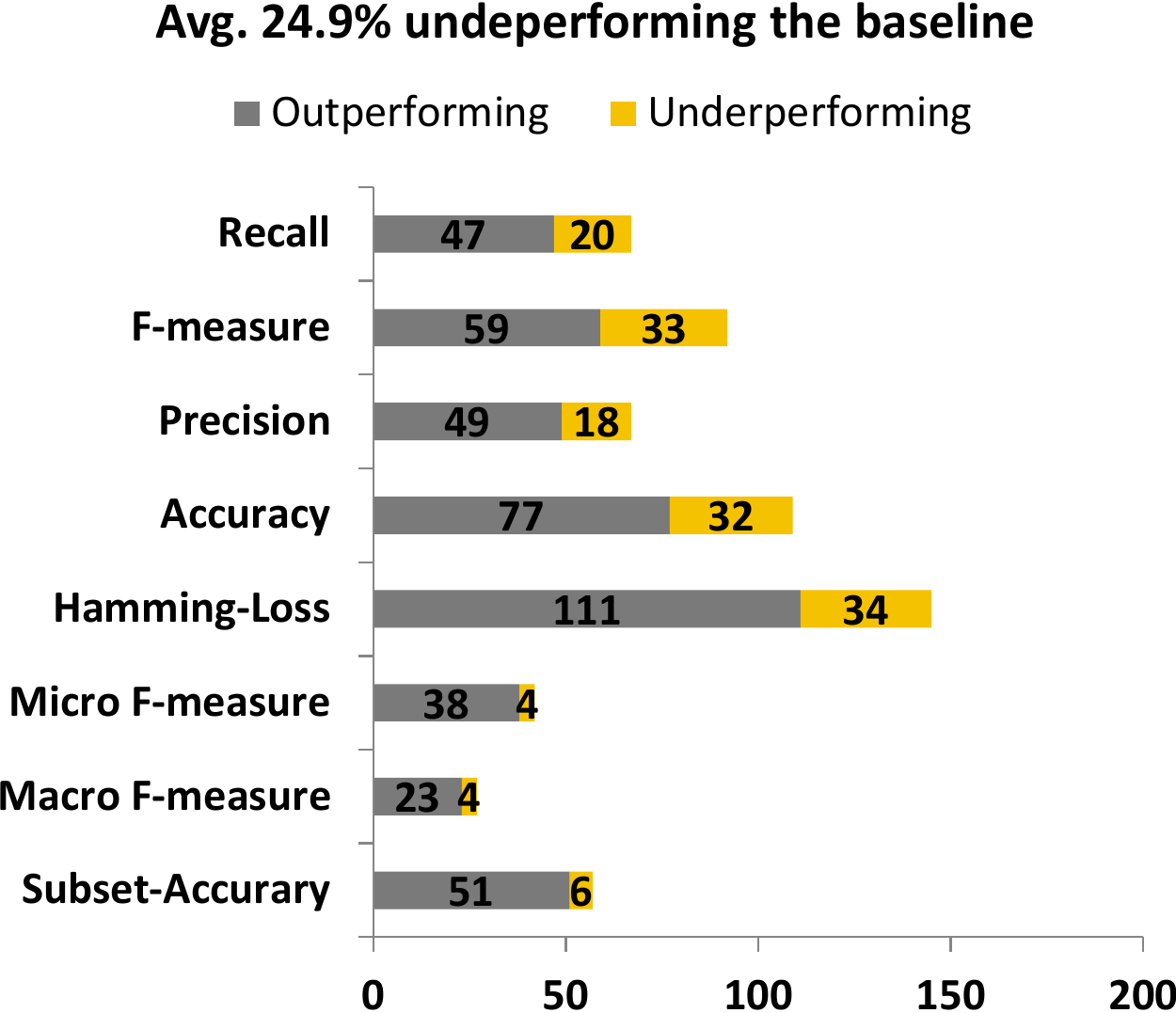}
} \subfigure[\emotions]{
    \includegraphics[width=0.4\textwidth]{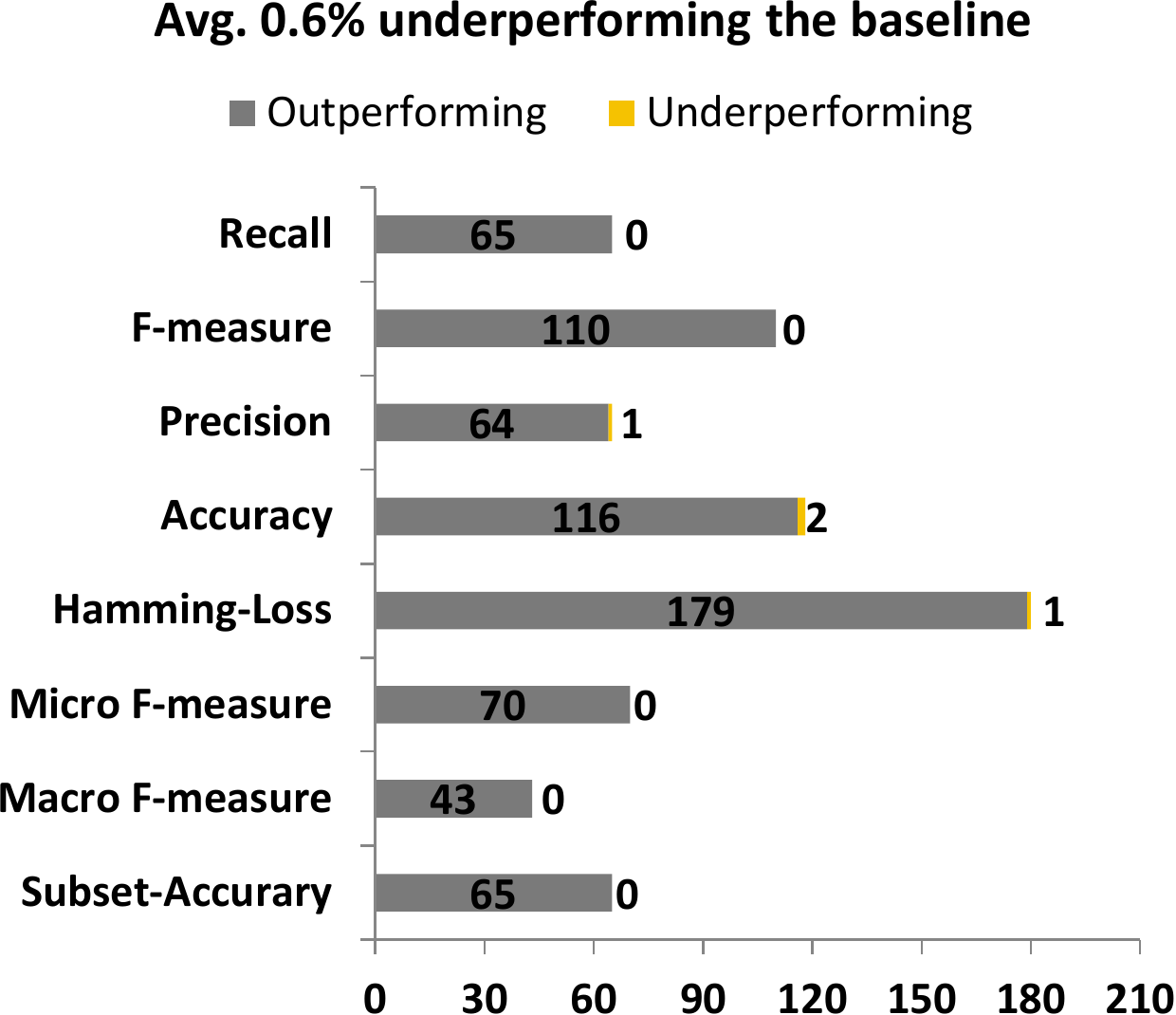}
}


\caption{Overall performance per measure of datasets
\corel, \mediamill, \enron and \emotions}
\label{fig:overallPerformance2}
\end{figure}

As can be observed, from a total of 5,342 measure values on the 10 datasets considered in
this work, 12.8\% are worse than or equal to the ones
provided by \lbg. Moreover, these worse results are concentrated
in some datasets, such as \corel, \mediamill and \enron, as shown in Figure~\ref{fig:overallPerformancepercent}.  On the
other hand, only 4 out of 716 (0.6\%) of the measure values published
for the \emotions dataset fulfill this condition.
Figure~\ref{fig:overallPerformance2} shows information of these
four datasets.

\begin{figure}[hp]
\centering 
\subfigure[{\scriptsize \corel~- \accuracy}]{
    \includegraphics[scale=0.32]{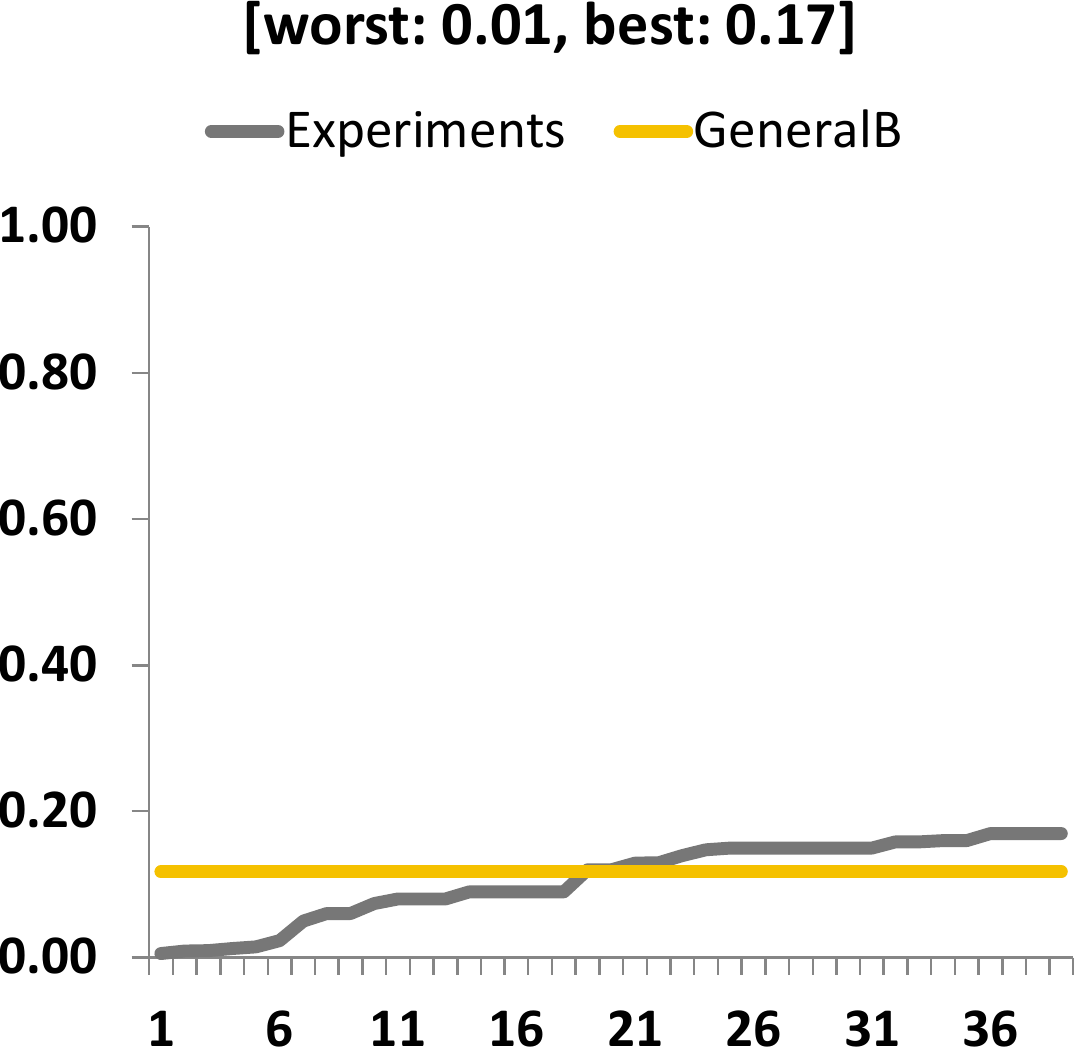}
}
\subfigure[{\scriptsize \corel~- \fmeasure}]{
    \includegraphics[scale=0.32]{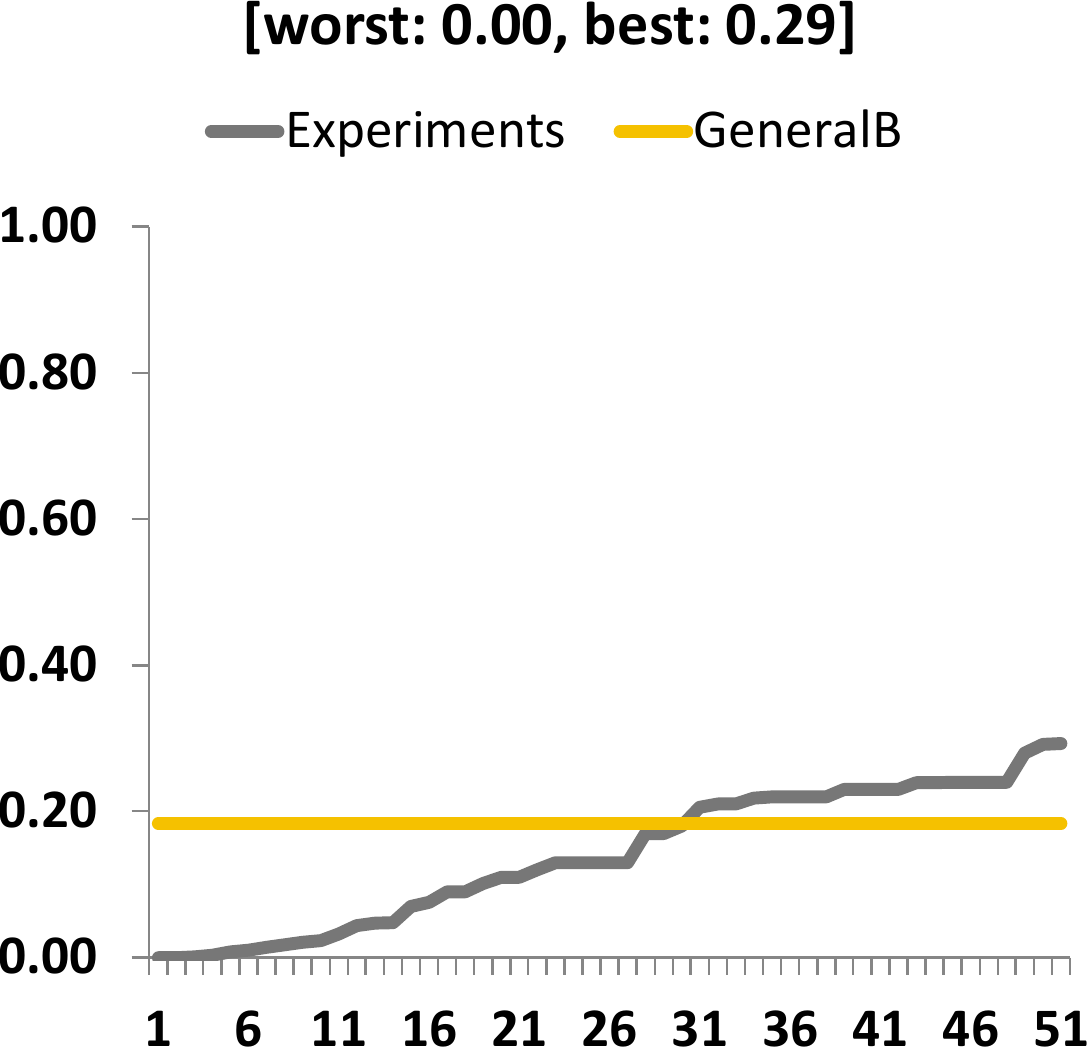}
}
\subfigure[{\scriptsize \corel~- \hloss}]{
    \includegraphics[scale=0.32]{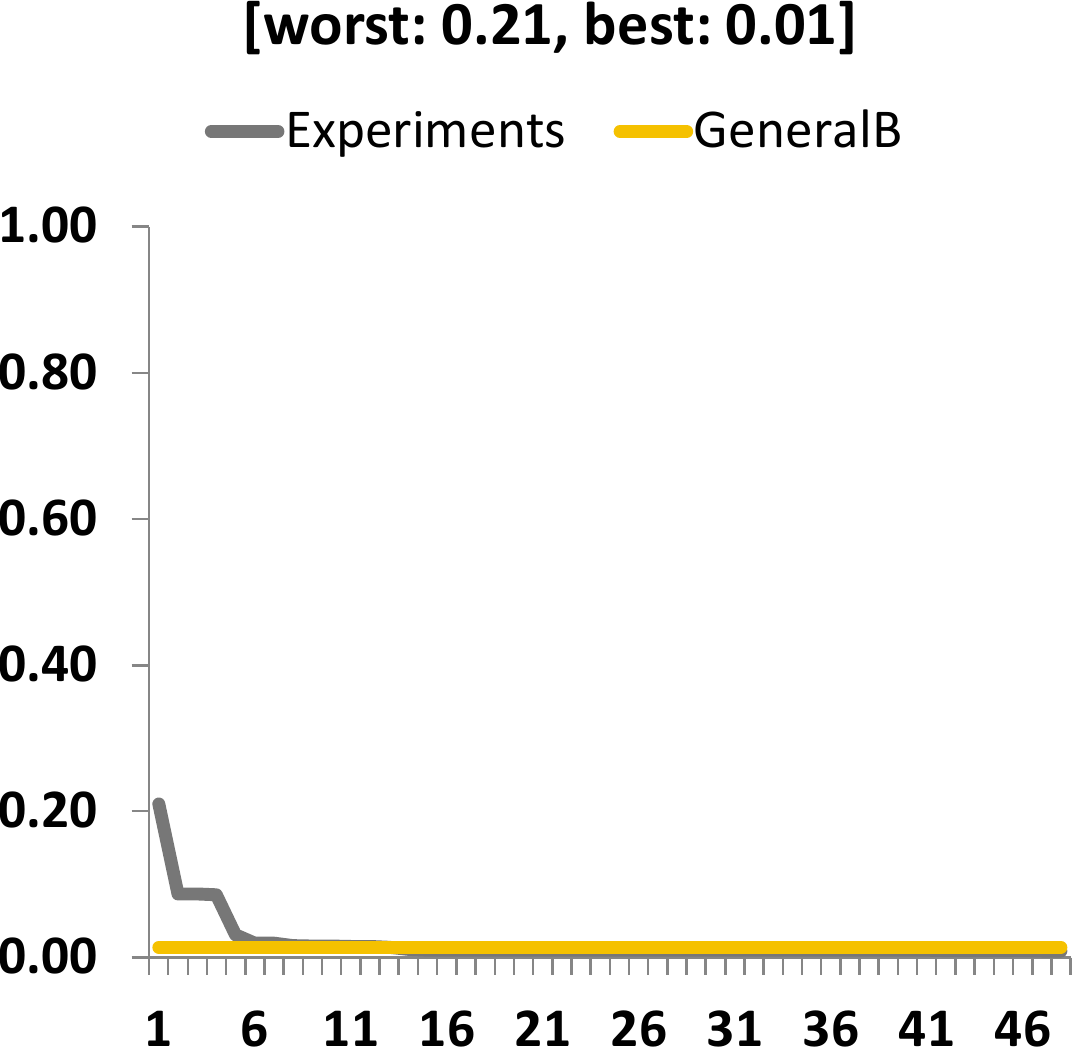}
}

\subfigure[{\scriptsize \mediamill~- \accuracy}]{
    \includegraphics[scale=0.32]{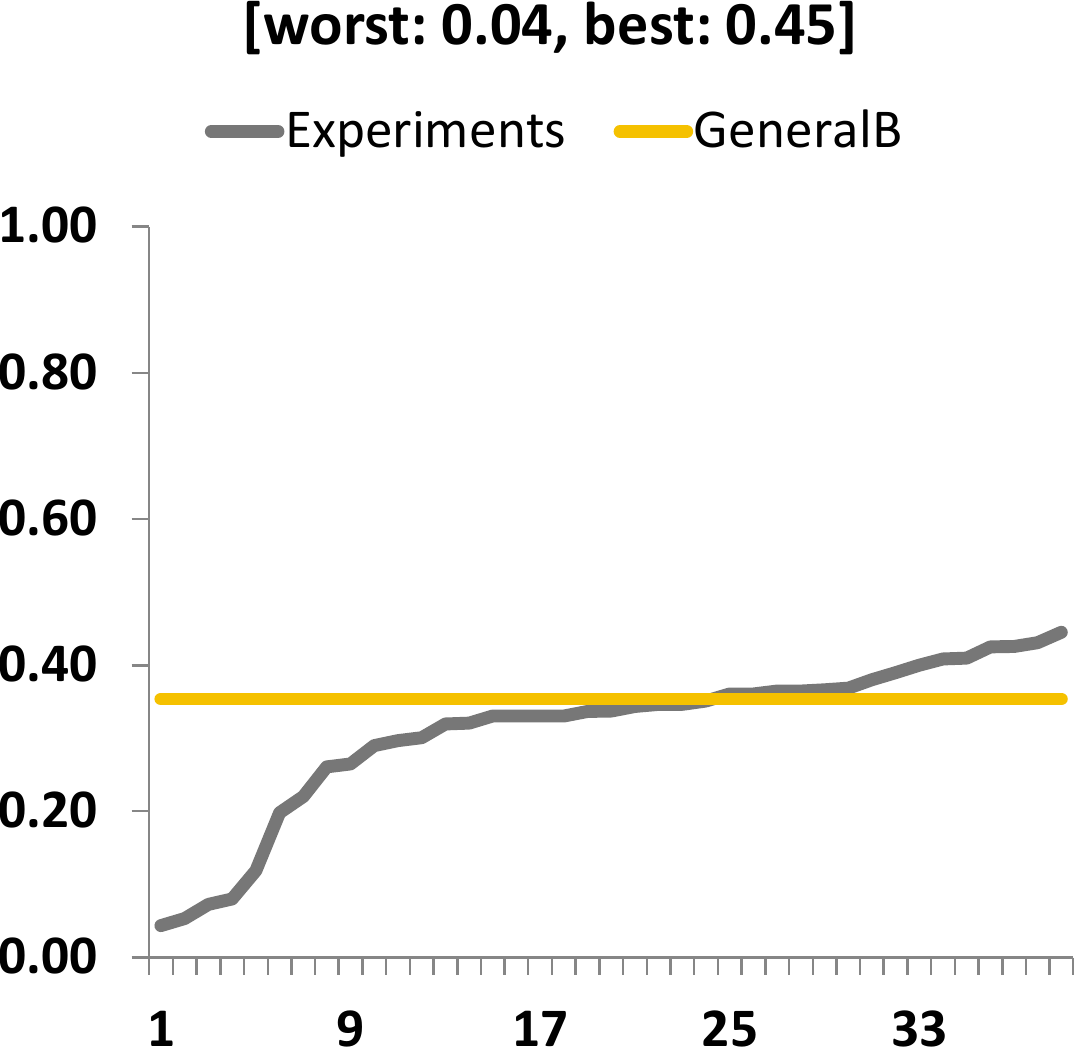}
}
\subfigure[{\scriptsize \mediamill~- \fmeasure}]{
    \includegraphics[scale=0.32]{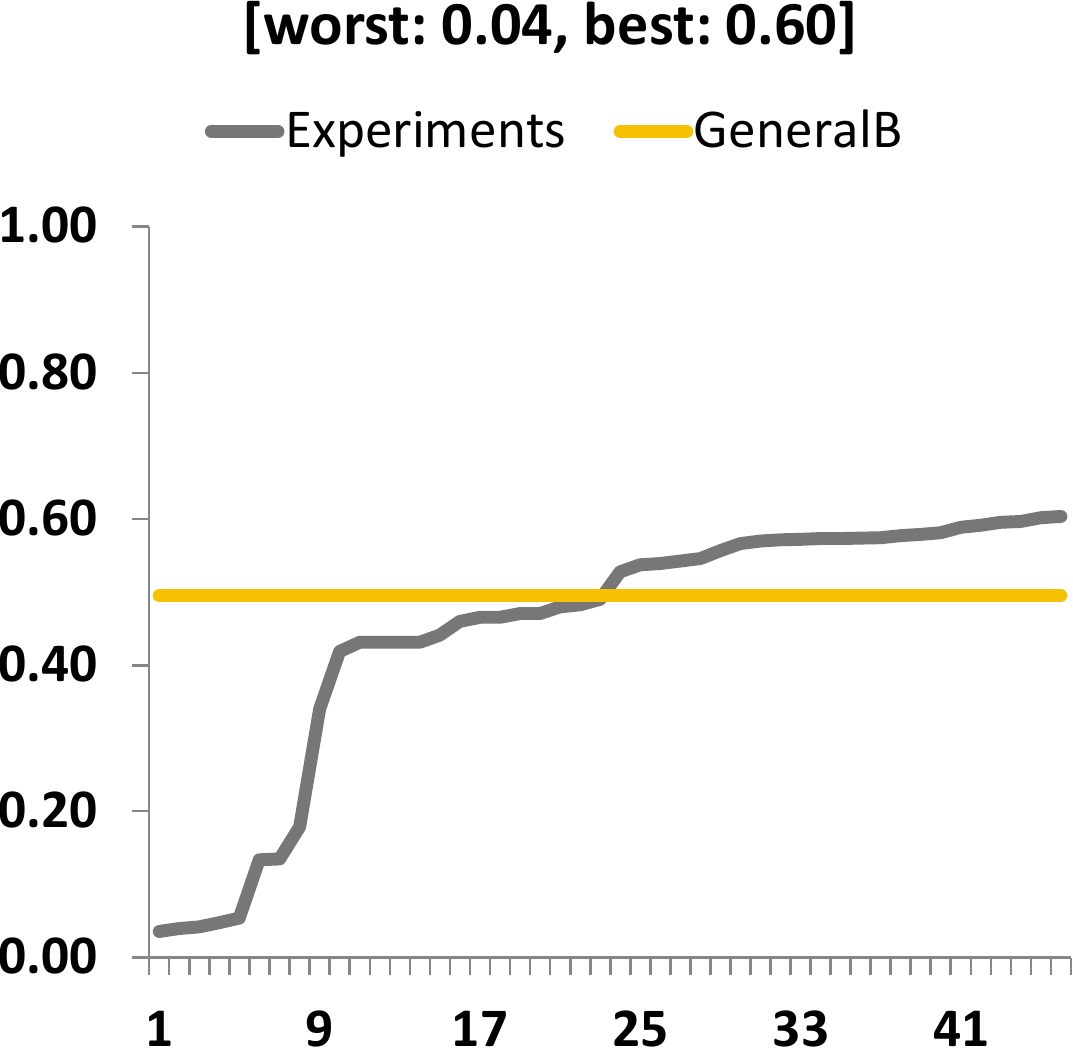}
}
\subfigure[{\scriptsize \mediamill~- \hloss}]{
    \includegraphics[scale=0.32]{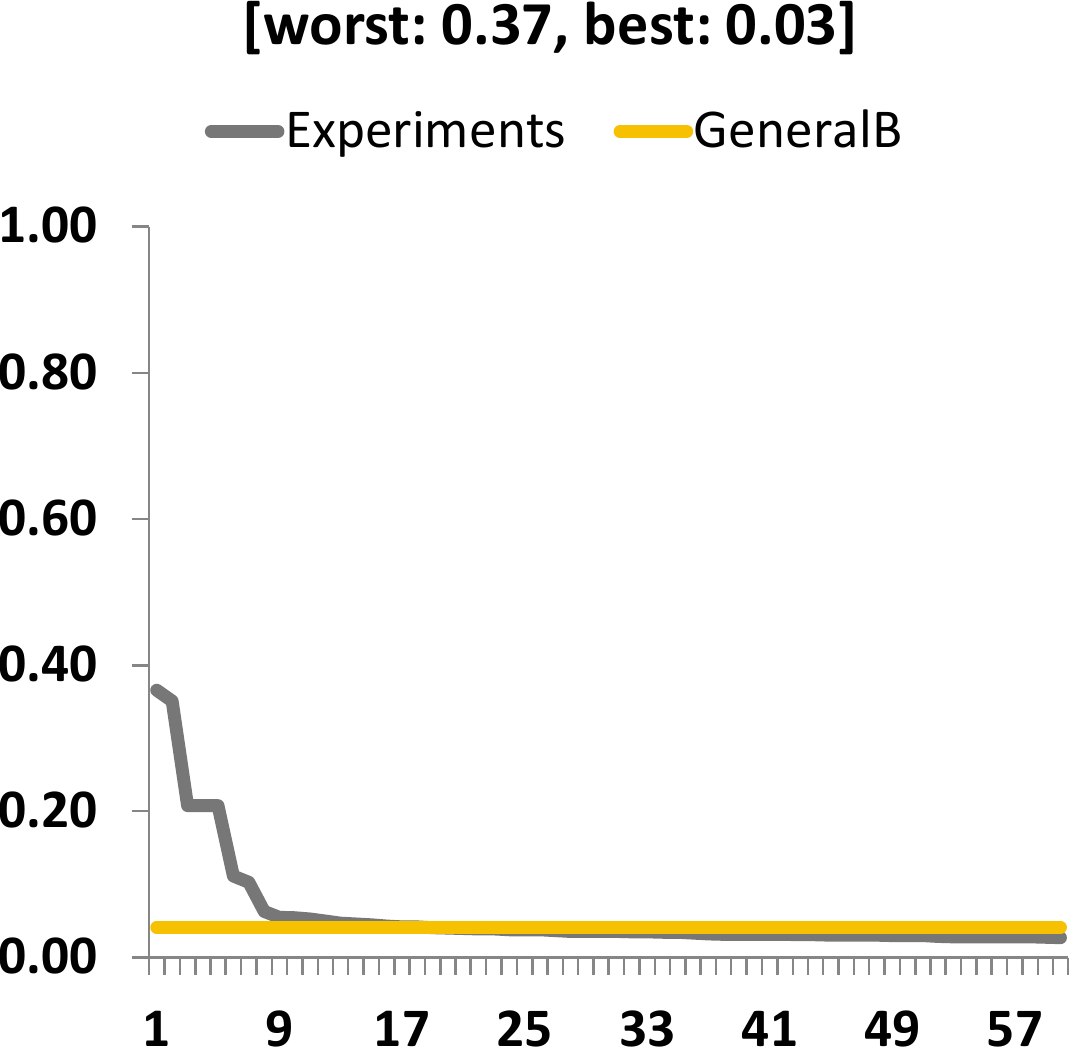}
}

\subfigure[{\scriptsize \enron~- \accuracy}]{
    \includegraphics[scale=0.32]{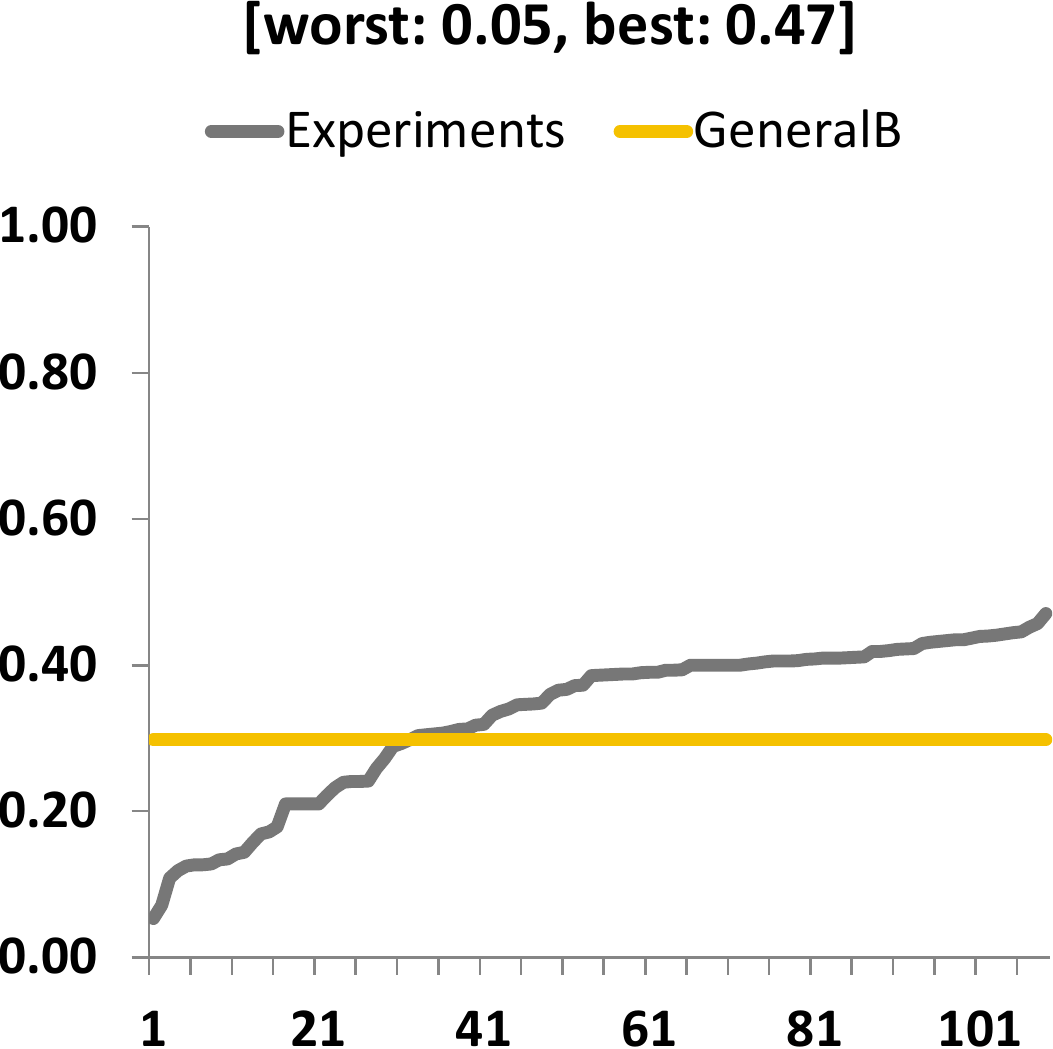}
}
\subfigure[{\scriptsize \enron~- \fmeasure}]{
    \includegraphics[scale=0.32]{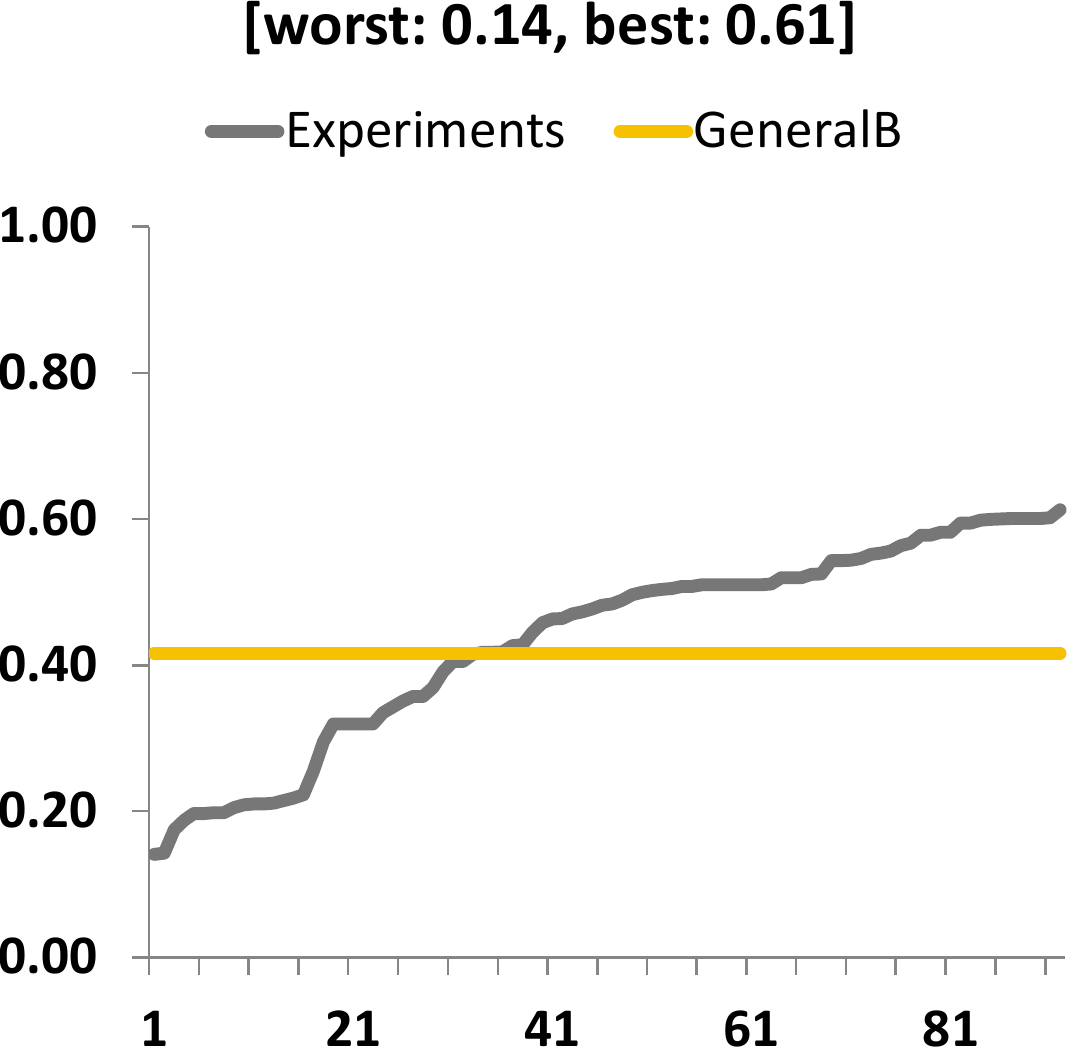}
}
\subfigure[{\scriptsize \enron~- \hloss}]{
    \includegraphics[scale=0.32]{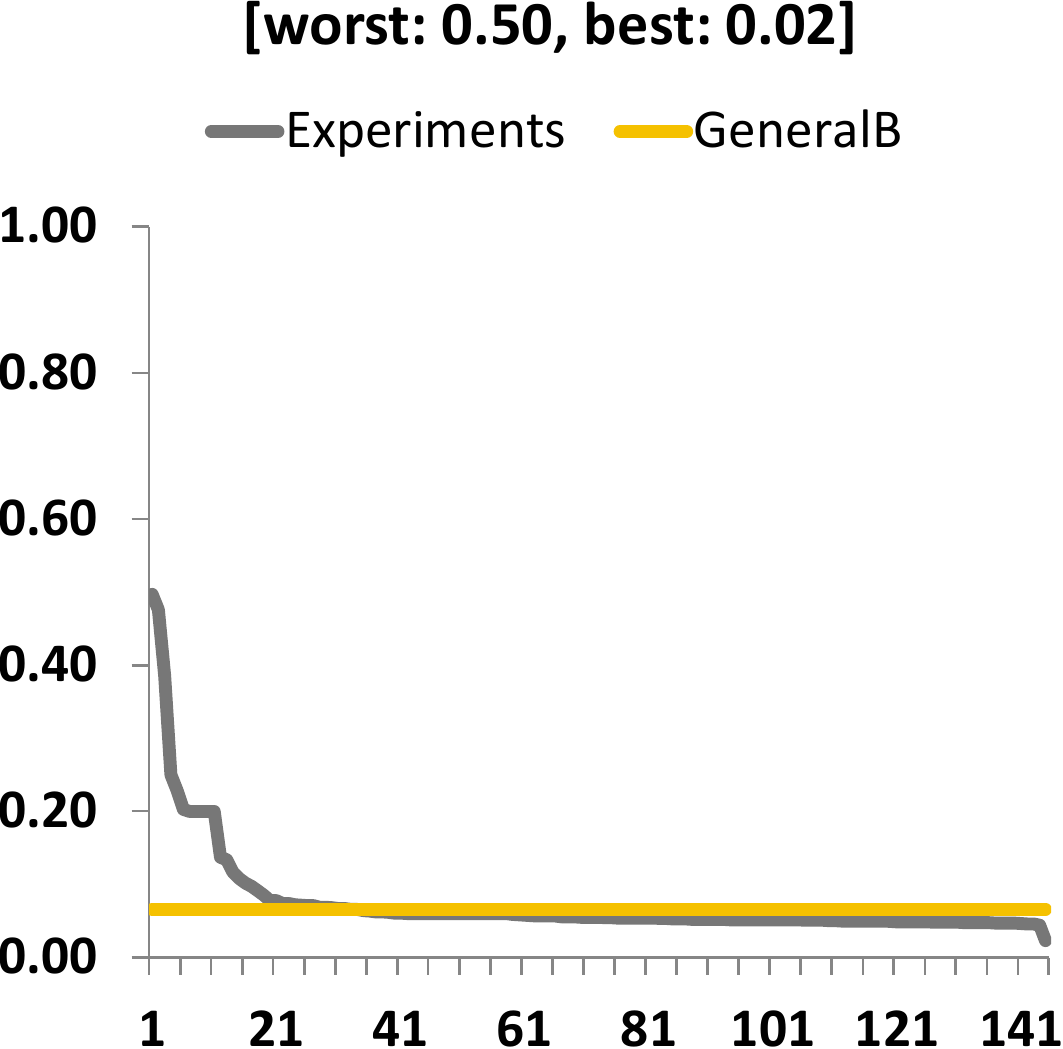}
}

\subfigure[{\scriptsize \emotions~- \accuracy}]{
    \includegraphics[scale=0.32]{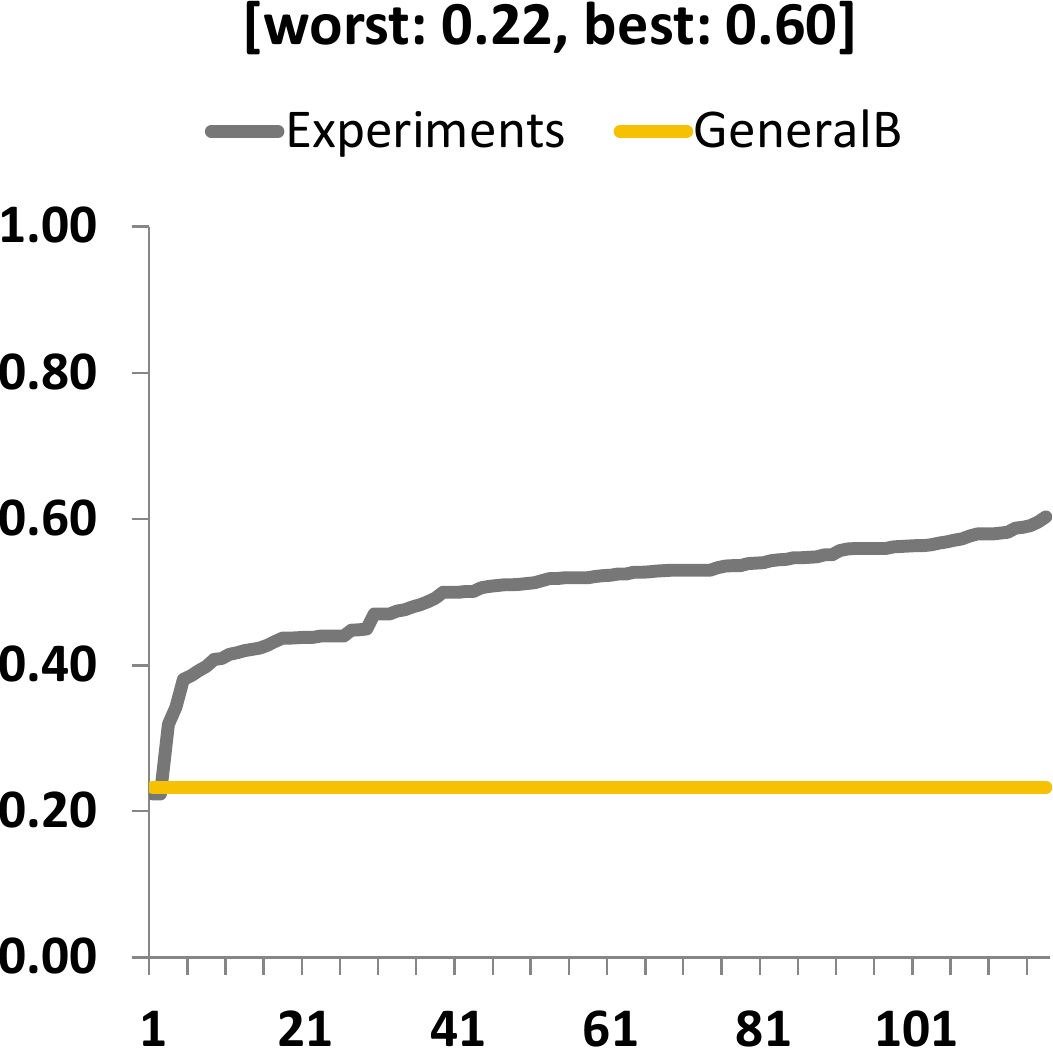}
}
\subfigure[{\scriptsize \emotions~- \fmeasure}]{
    \includegraphics[scale=0.32]{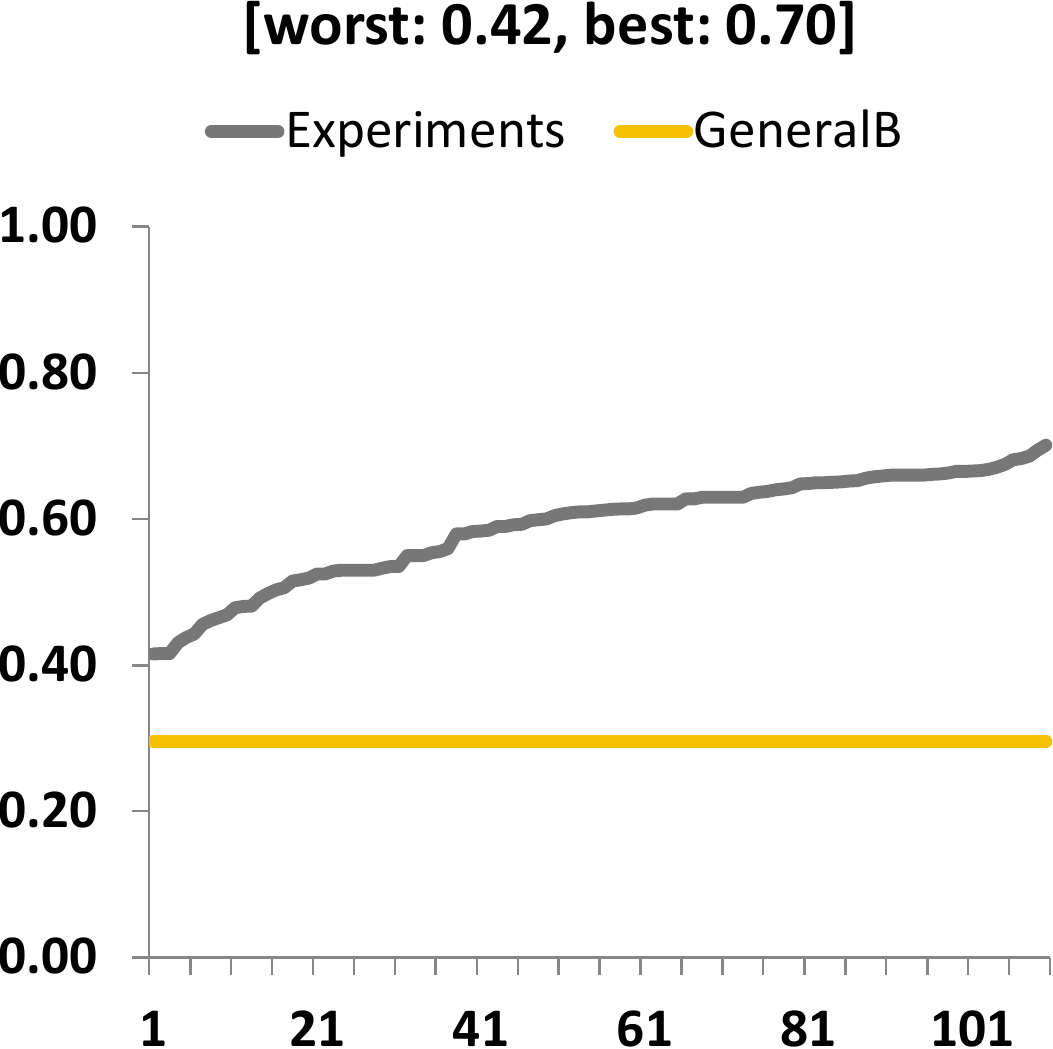}
}
\subfigure[{\scriptsize \emotions~- \hloss}]{
    \includegraphics[scale=0.32]{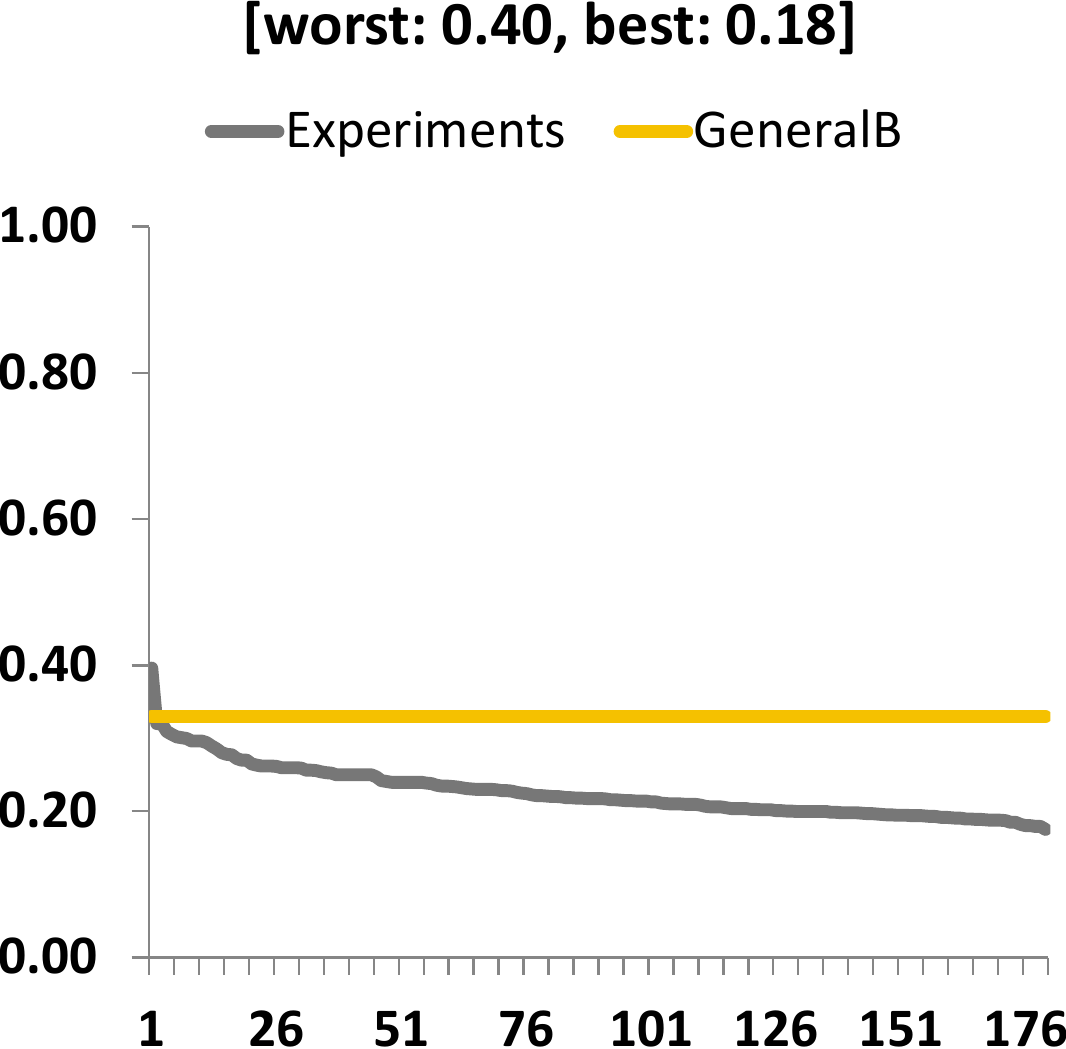}
}

%
%
%
%
\caption{Distribution of \accuracy, \fmeasure and \hloss
 evaluation measures values for datasets \corel, \mediamill, \enron and \emotions}

\label{fig:overallPerformance3}
\end{figure}

Nevertheless, this kind of information does not show the degree of
disagreement between the evaluation measure values published and the
ones provided by \lbg. To this end, we have extracted statistics from these values, as shown  in Figure~\ref{fig:overallPerformance3} for
the datasets \corel, \mediamill, \enron and \emotions  considering
the distribution of \accuracy, \fmeasure and \hloss measure values.
It also shows, in brackets, the worst and the best value found in the publications. 
Recall that for \hloss, the smaller the
value, the better the multi-label classifier performance is, while
for the others, greater values indicate better performance.

In fact, this sort of statistics extraction and organization was carried out for all datasets
and measures considered, and can be found at 
{\footnotesize\url{http://www.labic.icmc.usp.br/pub/mcmonard/ExperimentalResults/Metz-GeneralB-SupplementaryMaterial}}.

Figure~\ref{fig:overallPerformance3} shows that, in some cases, there is a
considerable gap between the worst and the best published measure 
values. Although this gap could be justified because different
multi-label algorithms minimize different loss-functions, which in
turn favors specific evaluation measures, it should be expected that special explanations are provided case these measures are worse than the ones from the simple
baseline classifier \lbg.

Furthermore, considering in Figure~\ref{fig:overallPerformance3} the measures
which are better than or equal to the ones from \lbg, it can be observed that
there is little improvement in those measures for \corel, \mediamill
and \enron datasets. On the other hand, the improvement is
considerable for \emotions. 

Table~\ref{tab:improvement} shows,  for
the 10 datasets, the highest ($\uparrow$) and the lowest
($\downarrow$) measure values published in the 64 papers for the 8
evaluation measures considered in this work, as well as the ones
from \lbg ($G_B$). Light gray cells indicate that the difference between 
the highest and the lowest measure values is greater than or equal to 0.5.
In most cases, it can be observed that there is a
very high discrepancy between the highest and the lowest published
measure values.

\begin{table}[h!]
  \centering
  \caption{Highest ($\uparrow$) and lowest ($\downarrow$) values reported for each dataset and the corresponding \lbg ($G_B$) values}
  \scriptsize
  \resizebox{\textwidth}{!}{
    \begin{tabular}{l ccc | ccc| ccc | ccc}

    \toprule
Dataset & $\uparrow$ & $\downarrow$ & $G_B$ & $\uparrow$ & $\downarrow$ & $G_B$ & $\uparrow$ & $\downarrow$ & $G_B$ & $\uparrow$ & $\downarrow$ & $G_B$ \\ 

\midrule

 & \multicolumn{3}{c}{\acc} &\multicolumn{3}{c}{\f} &\multicolumn{3}{c}{\hl} & \multicolumn{3}{c}{\pr} \\
 
\corel 				& 0.17 & 0.01 & 0.12 & 0.29 & 0.00 & 0.18 & 0.21 & 0.01 & 0.01 & \multicolumn{1}{g}{0.62} & \multicolumn{1}{g}{0.00} & 0.20 \\ 
\mediamill 		& 0.45 & 0.04 & 0.35 & \multicolumn{1}{g}{0.60} & \multicolumn{1}{g}{0.04} & 0.50 & 0.37 & 0.03 & 0.04 & \multicolumn{1}{g}{0.80} & \multicolumn{1}{g}{0.06} & 0.53 \\ 
\enron 			& 0.47 & 0.05 & 0.30 & 0.61 & 0.14 & 0.42 & 0.50 & 0.02 & 0.07 & \multicolumn{1}{g}{0.73} & \multicolumn{1}{g}{0.13} & 0.48 \\ 
\slashdot 		& \multicolumn{1}{g}{0.53} & \multicolumn{1}{g}{0.01} & 0.15 & \multicolumn{1}{g}{0.56} & \multicolumn{1}{g}{0.05} & 0.15 & \multicolumn{1}{g}{0.95} & \multicolumn{1}{g}{0.04} & 0.09 & \multicolumn{1}{g}{0.71} & \multicolumn{1}{g}{0.03} & 0.15 \\ 
\yeast 				& 0.57 & 0.33 & 0.42 & 0.69 & 0.33 & 0.55 & 0.30 & 0.08 & 0.26 & 0.75 & 0.34 & 0.58 \\ 
\bibtex 			& 0.38 & 0.01 & 0.07 & 0.46 & 0.03 & 0.10 & 0.21 & 0.01 & 0.03 & \multicolumn{1}{g}{0.64} & \multicolumn{1}{g}{0.02} & 0.11 \\ 
\genbase 		& \multicolumn{1}{g}{0.99} & \multicolumn{1}{g}{0.00} & 0.26 & \multicolumn{1}{g}{1.00} & \multicolumn{1}{g}{0.02} & 0.26 & \multicolumn{1}{g}{0.99} & \multicolumn{1}{g}{0.00} & 0.06 & \multicolumn{1}{g}{1.00} & \multicolumn{1}{g}{0.00} & 0.26 \\ 
\medical 		& \multicolumn{1}{g}{0.80} & \multicolumn{1}{g}{0.01} & 0.21 & \multicolumn{1}{g}{0.83} & \multicolumn{1}{g}{0.03} & 0.23 & \multicolumn{1}{g}{0.97} & \multicolumn{1}{g}{0.01} & 0.04 & \multicolumn{1}{g}{0.84} & \multicolumn{1}{g}{0.01} & 0.27 \\ 
\scene 			& \multicolumn{1}{g}{0.77} & \multicolumn{1}{g}{0.00} & 0.19 & \multicolumn{1}{g}{0.79} & \multicolumn{1}{g}{0.17} & 0.20 & 0.41 & 0.08 & 0.27 & \multicolumn{1}{g}{0.91} & \multicolumn{1}{g}{0.00} & 0.22 \\ 
\emotions 		& 0.60 & 0.22 & 0.23 & 0.70 & 0.42 & 0.30 & 0.40 & 0.18 & 0.33 & 0.74 & 0.43 & 0.45 \\ 

 \\ \midrule

  & \multicolumn{3}{c}{\re} &\multicolumn{3}{c}{\sacc} &\multicolumn{3}{c}{\fM} & \multicolumn{3}{c}{\fm} \\

\corel 				& \multicolumn{1}{g}{0.51} & \multicolumn{1}{g}{0.00} & 0.17 & 0.02 & 0.00 & 0.00 & 0.04 & 0.00 & 0.00 & 0.29 & 0.00 & 0.19 \\ 
\mediamill 		& \multicolumn{1}{g}{0.70} & \multicolumn{1}{g}{0.05} & 0.53 & 0.12 & 0.00 & 0.00 & 0.19 & 0.00 & 0.03 & \multicolumn{1}{g}{0.63} & \multicolumn{1}{g}{0.01} & 0.50 \\ 
\enron 			& \multicolumn{1}{g}{0.81} & \multicolumn{1}{g}{0.07} & 0.39 & 0.22 & 0.00 & 0.00 & 0.17 & 0.01 & 0.04 & 0.60 & 0.35 & 0.45 \\ 
\slashdot 		& \multicolumn{1}{g}{0.71} & \multicolumn{1}{g}{0.01} & 0.15 & 0.44 & 0.00 & 0.14 & 0.24 & 0.01 & 0.10 & 0.50 & 0.42 & 0.14 \\ 
\yeast 				& \multicolumn{1}{g}{0.82} & \multicolumn{1}{g}{0.32} & 0.55 & 0.28 & 0.04 & 0.05 & \multicolumn{1}{g}{0.87} & \multicolumn{1}{g}{0.03} & 0.21 & \multicolumn{1}{g}{0.85} & \multicolumn{1}{g}{0.04} & 0.57 \\ 
\bibtex 			& \multicolumn{1}{g}{0.65} & \multicolumn{1}{g}{0.05} & 0.11 & 0.27 & 0.06 & 0.00 & 0.32 & 0.05 & 0.00 & 0.46 & 0.12 & 0.00 \\ 
\genbase 		& \multicolumn{1}{g}{1.00} & \multicolumn{1}{g}{0.00} & 0.26 & 0.98 & 0.96 & 0.26 & 0.00 & 0.00 & 0.00 & 0.99 & 0.98 & 0.23 \\ 
\medical 		& \multicolumn{1}{g}{0.94} & \multicolumn{1}{g}{0.03} & 0.21 & \multicolumn{1}{g}{0.78} & \multicolumn{1}{g}{0.00} & 0.16 & 0.37 & 0.02 & 0.01 & 0.81 & 0.34 & 0.24 \\ 
\scene 			& \multicolumn{1}{g}{0.95} & \multicolumn{1}{g}{0.00} & 0.19 & \multicolumn{1}{g}{0.74} & \multicolumn{1}{g}{0.17} & 0.17 & 0.78 & 0.51 & 0.06 & 0.77 & 0.52 & 0.21 \\ 
\emotions 		& \multicolumn{1}{g}{0.79} & \multicolumn{1}{g}{0.28} & 0.23 & 0.35 & 0.08 & 0.07 & 0.73 & 0.37 & 0.10 & 0.73 & 0.44 & 0.31 \\ 
    \bottomrule
    \end{tabular}
    }
\label{tab:improvement}
\end{table}

Regarding the multi-label algorithms used in the 64 papers, most of
them follow the problem transformation approach,  using
state-of-the-art single-label learning algorithms as a base learner.
Binary Relevance is the most frequently used approach.

At this point, it is worth observing that we are quite confident
about the correctness (with respect to the published results) of the
collected measure values from the 64 papers. As stated earlier in
Section~\ref{sec:systematicreview}, these values were initially
double checked. After making the graphs for all datasets
and measures considered in this work, we checked, once more, the worst and the best published values.

From this third inspection of the gathered data, it was observed
that few papers explain and justify very poor results. However, similar to single-label learning, case the
multi-label community decides to adopt a simple baseline classifier
such as \lbg, or any other, we think that it will encourage
the authors to provide special explanations on very poor results.

\section{Conclusions and Future Work} \label{sec:conclusion}

The single-label community expects that in non skewed domains a simple baseline classifier, which always predicts the majority class, should do worse than classifiers constructed by a learning algorithms. However, to the best of our knowledge, the multi-label community still does not have a consolidated idea of a simple multi-label baseline classifier.

Aiming to raise awareness  of considering a simple multi-label baseline classifier, we have carried out a systematic review of the multi-label learning literature in order to collect experimental results to contrast with the proposed simple multi-label baseline classifier \lbg. 

It was found that an important number of published results (12.8\%) are worse than or equal to the ones obtained by \lbg. In fact, for all the 10 most frequently used datasets presented in the work, results worse than or equal to the ones obtained by \lbg were found. In the extreme case, 43\% of the published results for one dataset are worse than or equal to the \lbg results.

Although we do not claim that the proposed \lbg multi-label baseline classifier should be the one to be used by the community, we hope that this work would encourage the multi-label community to consider the idea of using a simple baseline classifier as an initial reference related to the learning power of multi-label algorithms. With the use of a baseline, built by only taking into account the label distribution information, it would be possible to identify cases where the obtained results are not reasonable enough, and give support for  better explanations about these results.

As future work, we plan to increase the number of electronic databases to search for publications which answer our research question and do not fulfil any of the exclusion criteria. As the organization of the information extracted allows to answer several useful questions, such as  \textit{Which  publications use algorithm A on dataset B using 10-fold cross-validation and what are the results obtained?} \textit{Are there publications reporting results on datasets with cardinality greater than C and a distinct number of multi-labels greater than W?}, we plan to increment and further structure the gathered information making it available to the community on a Web page.




\bibliographystyle{elsarticle-harv}

\end{document}